\documentclass[journal]{IEEEtran}
\usepackage[linesnumbered,lined,noend,ruled]{algorithm2e}
\usepackage{algorithmic}
\usepackage{amsmath,amssymb}
\usepackage{subfigure}
\usepackage{amstext}
\usepackage{amsfonts}
\usepackage{url}
\usepackage{bbm}
\usepackage{color}
\usepackage{float}

\usepackage{color}
\definecolor{red}{RGB}{255,0,0}

\usepackage{multirow}
\usepackage[export]{adjustbox}
\usepackage{ragged2e}
\usepackage{array}

\usepackage{xspace}
\newcommand{\ie}{\textit{i.e.}\xspace}
\newcommand{\eg}{\textit{e.g.}\xspace}
\newcommand{\etal}{\textit{et al.}\xspace}

\begin{document}
%
\title{Kernel Proposal Network for Arbitrary Shape \\ Text Detection}

\author{Shi-Xue Zhang$ ^{\textbf{*}} $, Xiaobin Zhu$ ^{\dagger} $, Jie-Bo Hou$ ^{\textbf{*}} $, Chun Yang, Xu-Cheng Yin,~\IEEEmembership{Senior Member,~IEEE}
\thanks{S. Zhang, X. Zhu, J. Hou, and C. Yang are with the School of Computer and Communication Engineering, University of Science and Technology Beijing (USTB), Beijing, 100083, China, e-mail: zhangshixue111@163.com; houjiebo@gmail.com; \{zhuxiaobin, chunyang\}@ustb.edu.cn.}
\thanks{X. Yin  is with the School of Computer and Communication Engineering, and Institute of Artificial Intelligence, University of Science and Technology Beijing (USTB), Beijing, 100083, China, also with USTB-EEasyTech Joint Lab of Artificial Intelligence, Beijing, 100083, China, e-mail: xuchengyin@ustb.edu.cn.}
\thanks{$ ^{\dagger} $: Corresponding authors ; *Authors contributed equally.}
\thanks{Manuscript received July 5, 2021; revised January 15, 2022.}}

%
%

\markboth{Journal of \LaTeX\ Class Files,~Vol.~14, No.~8, January~2022}%
{Shell \MakeLowercase{\textit{et al.}}: Bare Demo of IEEEtran.cls for IEEE Journals}
%

\maketitle

\begin{abstract}
Segmentation-based methods have achieved great success for arbitrary shape text detection. However, separating neighboring text instances is still one of the most challenging problems due to the complexity of texts in scene images. In this paper, we propose an innovative Kernel Proposal Network (dubbed KPN) for arbitrary shape text detection. The proposed KPN can separate neighboring text instances by classifying different texts into instance-independent feature maps, meanwhile avoiding the complex aggregation process existing in segmentation-based arbitrary shape text detection methods. To be concrete, our KPN will predict a Gaussian center map for each text image, which will be used to extract a series of candidate kernel proposals (i.e., dynamic convolution kernel) from the embedding feature maps according to their corresponding keypoint positions. To enforce the independence between kernel proposals, we propose a novel orthogonal learning loss (OLL) via orthogonal constraints. Specifically, our kernel proposals contain important self-information learned by network and location information by position embedding. Finally, kernel proposals will individually convolve all embedding feature maps for generating individual embedded maps of text instances. In this way, our KPN can effectively separate neighboring text instances and improve the robustness against unclear boundaries. To our knowledge, our work is the first to introduce the dynamic convolution kernel strategy to efficiently and effectively tackle the adhesion problem of neighboring text instances in text detection. Experimental results on challenging datasets verify the impressive performance and efficiency of our method. The code and model are available at https://github.com/GXYM/KPN.		
\end{abstract}


\IEEEpeerreviewmaketitle

\section{Introduction}
\label{intro}
\begin{figure}[htbp]
	\subfigure[FCN-based]{
		\begin{minipage}[t]{0.431\linewidth}
			\centering
			\includegraphics[width=0.98\linewidth]{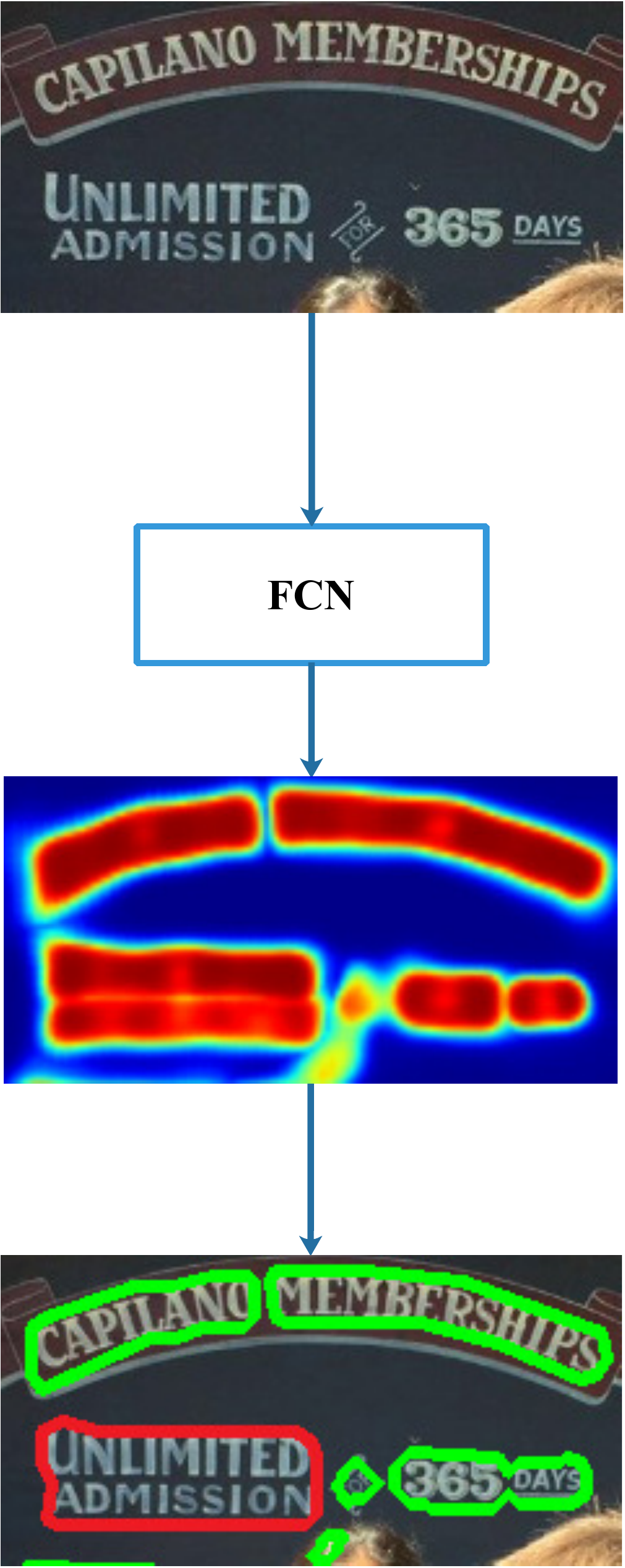}
		\end{minipage}%
	}
	\subfigure[KPN]{
		\begin{minipage}[t]{0.545\linewidth}
			\centering
			\includegraphics[width=0.98\linewidth]{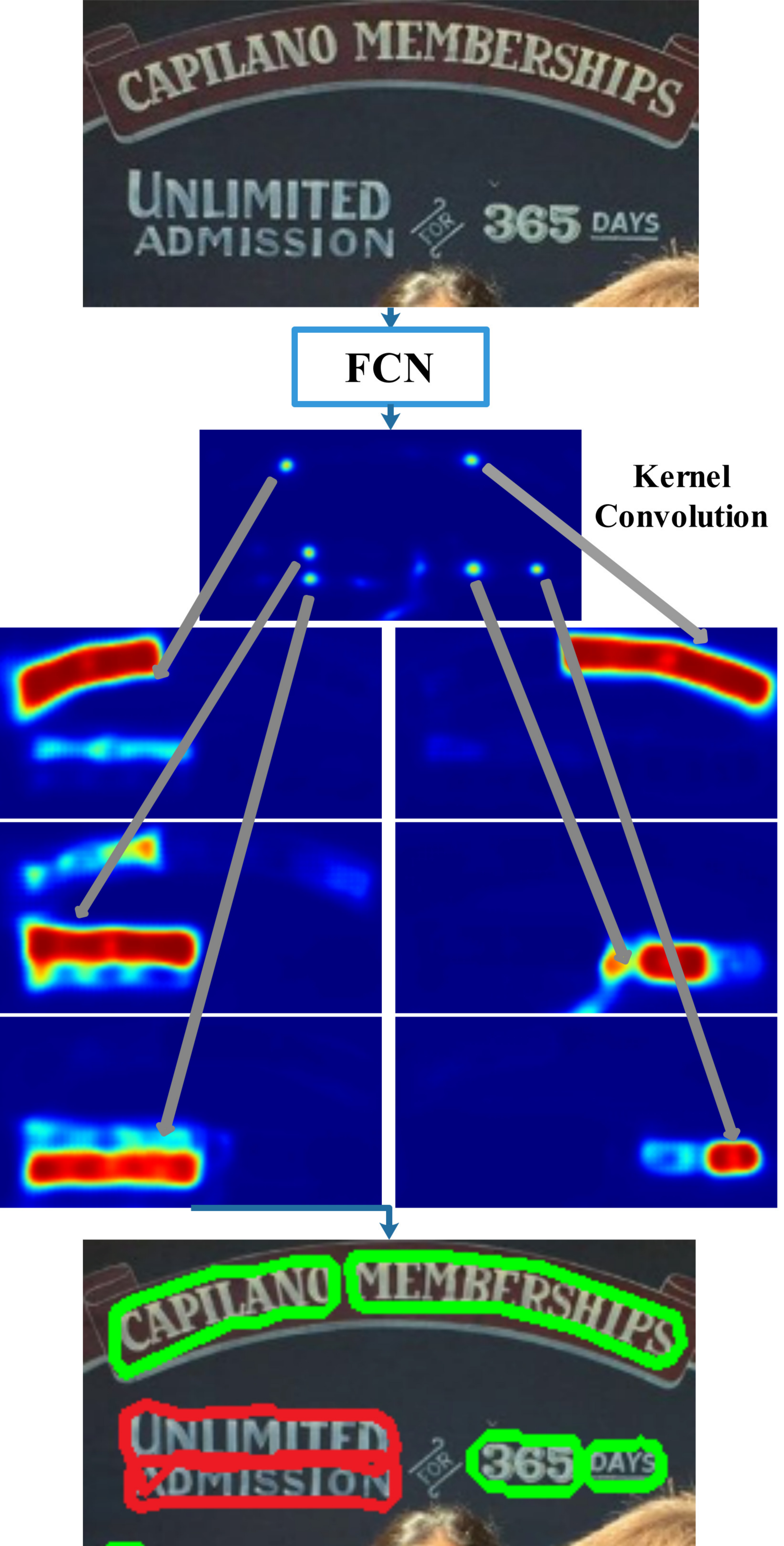}
		\end{minipage}%
	}
\caption{Comparison with FCN-based methods. (a) shows FCN-based methods suffer from covering adjacent instances (red contour in the bottom image). (b) demonstrates that our KPN can successfully disperse texts on feature maps, thus generating separated yet accurate masks for text instances.}\label{fig:introduction}
\end{figure}
\IEEEPARstart{S}{cene} text detection is a fundamental and critical task in computer vision because it is a key step in various text-related applications, including translation, text-visual question answering, text recognition, text mining.  With the rapid development of deep learning-based object detection~\cite{Faster-rcnn,FPN, DABNet} and segmentation~\cite{FCN,U-Net},  scene text detection has witnessed great progress~\cite{EAST,RRPN,CVPR2020_DRRG}. Arbitrary shape scene text detection, as one of the most challenging tasks in text detection, has attracted ever-increasing interest in both research and industrial communities. Except for the challenges existing in the general scene text detection tasks, arbitrary shape text detection should address additional challenging problems, such as varied scales, curved and arbitrary shapes.

Recently, segmentation-based methods~\cite{PixelLink,PAN,CVPR19_Embedding} have achieved promising performance in detecting arbitrary shape text. Benefiting from the adaptability of pixel-level prediction (\eg, pixel-wise text/non-text classification mask), segmentation-based methods~\cite{DB,PAN,CVPR19_PSENet} based on fully convolutional neural networks (FCNs) can easily adapt to irregular texts. However, segmentation-based text detection methods always suffer from separating neighboring text instances in complex scene images, as shown at the top of Fig. \ref{fig:introduction}(a), which may be caused by inaccurate annotations or similar appearances between neighboring text instances. To solve this problem, many segmentation-based methods~\cite{PixelLink,PAN,CVPR19_Embedding,CVPR19_PSENet,TextField} adopt complicated post-processing operations to group and separate text instances based on their predicted masks. PSENet~\cite{CVPR19_PSENet} adopts kernels with different scales for each text instance and gradually expands the minimal scale kernel to reconstruct individual text instances. The other methods~\cite{PixelLink,PAN,CVPR19_Embedding,TextField} try to learn pixel-pair embedding vectors for clustering text
pixels into different text instances during post-processing. Although the adhesion problem can be alleviated to some extent in~\cite{PixelLink,PAN,CVPR19_Embedding,CVPR19_PSENet,TextField}, they all suffer from the formidable computational cost in complex post-processing. To improve model efficiency, DB~\cite{DB} abandons complex post-processing and shrinks annotated boundaries with Vatti clipping algorithm~\cite{poly_clip} to obtain probability maps. Then, the probability maps will be used to restore separated boundaries for text instances by the inverse transformation of the Vatti clipping algorithm. Unfortunately, the text boundaries generated by fixed scaling rules in DB~\cite{DB} tend to be inaccurate on text with different scales. Besides, some Mask R-CNN \cite{Maskrcnn} based methods~\cite{FTSN,CVPR19_LOMO,Mask_TextSpotter_v3} firstly detect bounding rectangles of candidate texts, followed by pixel-wisely segmenting text instances in their corresponding boxes. Although Mask R-CNN based methods can alleviate the false detection of covering adjacent instances, their performance greatly relies on the extracted bounding boxes' quality and computational-cost RoI operation.

In this paper, we propose an innovative Kernel Proposal Network (dubbed KPN) for arbitrary shape text detection. The proposed KPN can separate neighboring text instances by classifying different texts into instance-independent feature maps, meanwhile avoiding the complex aggregation process existing in segmentation-based arbitrary shape text detection methods. To be concrete, our KPN will predict a Gaussian center map for each text image, which will be used to extract a series of candidate kernel proposals (\ie, dynamic convolution kernel) from the embedding feature maps according to their corresponding keypoint positions. However, the premise for accurately separating neighboring text is that these kernel proposals should be independent. To enforce the independence between kernel proposals, we propose a novel orthogonal learning loss (OLL) via orthogonal constraints.  Specifically,  our kernel proposals mainly contain important self-information learned by the network and position information by position embedding, instead of the shared information of all texts. After removing the noise kernel proposals via pre-defined rules, the remaining ones will individually convolve all embedding feature maps for classifying individual texts into instance-independent maps. In this way, our KPN can effectively separate neighboring text instances and improve the robustness against unclear boundaries. To our knowledge, our work is the first to introduce the dynamic convolution kernel strategy to efficiently and effectively tackle the adhesion problem of neighboring text instances in text detection. 

In summary, our main contributions are four-fold:

\begin{itemize}
	\medskip
	\item We develop a post-processing-free segmentation-based arbitrary shape text detection framework that cleverly avoids computational-cost post-processing for achieving effectiveness and efficiency.
	
	\medskip
	\item We propose a novel dynamic convolution kernel strategy for text detection in which kernel proposals contain important self-information and position information, resulting in the effectiveness of separating neighboring texts and improving the robustness against tiny intervals or unclear boundaries.
	
	\medskip
	\item  We propose a novel orthogonal learning loss (OLL) that directly enforces the independence between kernel proposals via orthogonal constraints.
	
	\medskip
	\item Experiments conducted on publicly available datasets demonstrate the effectiveness and efficiency of the proposed method.
	
\end{itemize}

The rest of the paper is organized as follows: Section \ref{Related_Work} overviews the related work. Section \ref{Proposed_Method} elaborates our method. In Section \ref{Experiments}, we demonstrate experimental results on several datasets. Finally, we conclude our work in Section \ref{Conclusion}.

\section{Related Work} \label{Related_Work}
\subsection{Regression-based Methods} 
Regression-based methods generally predict text boxes by regressing offsets of bounding rectangles based on pre-defined anchors or original pixels, which can be roughly divided into two categories: anchor-based methods and anchor-free methods. Most of the Anchor-based methods try to design or learn appropriate anchors for accurately regressing text boxes. Both TextBoxes~\cite{textboxes} and TextBoxes++ \cite{textboxes++} adopt a series of anchors with different aspect ratios for covering texts with varied lengths. Specifically, TextBoxes++ regresses offsets of quadrilateral four points by pre-defined anchors for adapting to arbitrarily oriented texts. RRPN \cite{RRPN} adopts rotated anchors (3 scales, 3 ratios, and 6 angles) and Rotated RoI pooling (RRoI
pooling) for arbitrarily oriented text detection. Anchor-free methods \cite{EAST,DDR} try to regress texts without pre-defined anchors for improving model adaptability. As a classical anchor-free method, EAST \cite{EAST} adopts an FCN \cite{FCN} branch for classification, and a regression branch for directly regressing bounding quadrilateral via IoU loss \cite{IOU}. HAM \cite{HAM} proposes a Hidden Anchor Mechanism to integrate the advantages of the anchor-based method into the anchor-free method. However, the regression distance and angle in most regression-based methods are strictly confined to quadrilateral text, making it challenging to handle arbitrary shapes.

\subsection{Connected Component-based Methods} 
Component-based detection~\cite{PBNet, CTPN} is also an important direction in the field of object detection. In scene text detection, Connected Component-based (CC-based) methods~\cite{CTPN, SegLink, SegLink++, CRAFT, TextDragon, CVPR2020_DRRG} usually detect individual text parts or characters firstly, followed by post-processing of link or group for generating final texts. CTPN~\cite{CTPN} modifies the framework of Faster R-CNN~\cite{Faster-rcnn} to extract horizontal text components
of fixed width for connecting dense text components and generating horizontal text lines. SegLink~\cite{SegLink} decomposes each text into two detectable elements, \ie, segment and link, where a link indicates that a pair of adjacent segments belong to the same text. CRAFT \cite{CRAFT} detects text regions by exploring affinities between characters. TextDragon \cite{TextDragon} first detects local bounding boxes of texts and then groups them into different text instances via their geometric relations. Zhang \etal~\cite{CVPR2020_DRRG} 
adopted a graph convolution neural network (GCN) to learn and infer linking relationships between different text components for grouping them into final text instances. Although CC-based methods can learn a flexible representation for adapting to arbitrary shape text, the complex post-processing for grouping text components is always time-consuming.

\subsection{Contour-based Methods} 
Contour-based~\cite{CVPR19_ATRR, ContourNet, ABCNet, FCENet, TextRay, PCR, TextBPN} try to directly model the text boundary for detecting the arbitrary shape text. ABCNet~\cite{ABCNet} and FCENet~\cite{FCENet} model contours of text instances with curve modeling (Bezier-Curve and Fourier-Curve), which can well fit closed contour with progressive approximation. TextRay~\cite{TextRay} formulates text contours in the polar
system and proposes a single-shot anchor-free framework to predict geometric parameters and output simple polygon detections. PCR~\cite{PCR} proposes a progressive contour regression framework to detect arbitrary-shape scene texts. TextBPN~\cite{TextBPN} proposes an adaptive boundary deformation model to perform boundary deformation iteratively. However, compared with segmentation-based methods, 
the performance and efficiency of the contour-based methods are unsatisfactory.

\subsection{Segmentation-based Methods} 
Segmentation-based methods can get the contours of text instances from segmentation masks. In \cite{FTSN, IncepText, Mask_TextSpotter_v3}, they first predict the bounding boxes of texts and then segment the pixel-wise mask of text in each box. Some other methods~\cite{PixelLink, TextField, CVPR19_Embedding} utilize FCN \cite{FCN} to predict text masks and predict extra embedding vectors to cluster pixels in text masks. PixelLink \cite{PixelLink} predicts eight linkages to judge the connectivity and then link pixels using a disjoint-set data structure. TextField \cite{TextField} adopts a direction field to group pixels with morphological post-processing. PSENet \cite{CVPR19_PSENet} shrinks the text region into various scales for generating more distinct boundaries and then gradually expands the minimal scale kernel to the text instance with the complete shape. Tian \etal~\cite{CVPR19_Embedding} assumed each text instance as a cluster and predicted an embedding map via pixel clustering. Overall, segmentation-based methods can easily adapt to texts in arbitrary shapes. However, the existing methods still struggle with high complex post-processing to cluster pixels into texts instances.

\begin{figure}[htbp]
	\begin{center}
		\includegraphics[width=0.99\linewidth]{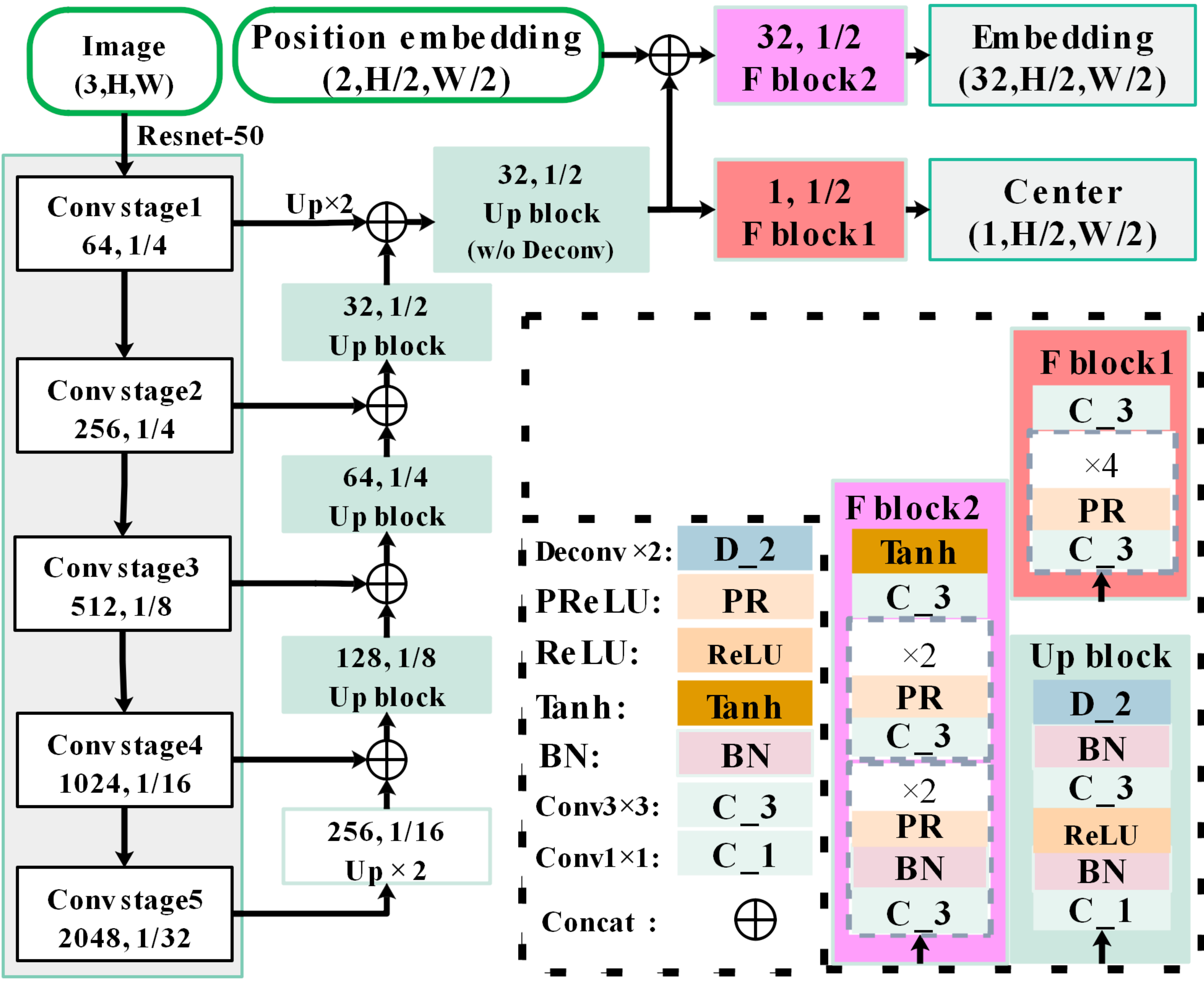}
	\end{center}
	\caption{Feature extraction sub-network has two output branches: 1 channel for center map and the other 32 channels for embedding feature maps.}
	\label{fig:network}
\end{figure}

\subsection{Dynamic Kernel Methods}
In the  instance segmentation task,  
AdaptIS \cite{AdaptIS} iteratively predict point proposals to generate instances. At each iteration, AdaptIS picks one point proposal and then uses the AdaIN mechanism \cite{AdaIN} to generate a corresponding instance. By providing different parameters to AdaIN, AdaptIS can vary the network output for the same input. However, AdaptIS predicts instances in inference iteratively with low efficiency. Based on FCOS~\cite{FCOS}, CondInst~\cite{condinst} use the conditional convolution (\ie dynamic convolution) to generate instance-sensitive filters to encode object instance information. SOLOv2 \cite{solov2} generates and separates instances simultaneously. More specifically, it dynamically proposes $s \times s$ (the input image is divided into $s \times s$ grids) kernels to convolve the feature maps for generating $s \times s$ instances in $s \times s$ channels. So, if we set the ``resolution" ($s \times s$) of kernels in SOLOv2 to $w \times h$ (width and height of feature maps) as the same as the general ``resolution" in scene test detection, the memory cost of SOLOv2 in feature maps will be $(w  \times h)^2$ which is computation and memory formidable. Moreover, CondInst and SOLOv2 still rely on time-consuming NMS for removing abundant duplicate boxes.

\begin{figure*}[htbp]
	\begin{center}
		\includegraphics[width=0.99\linewidth]{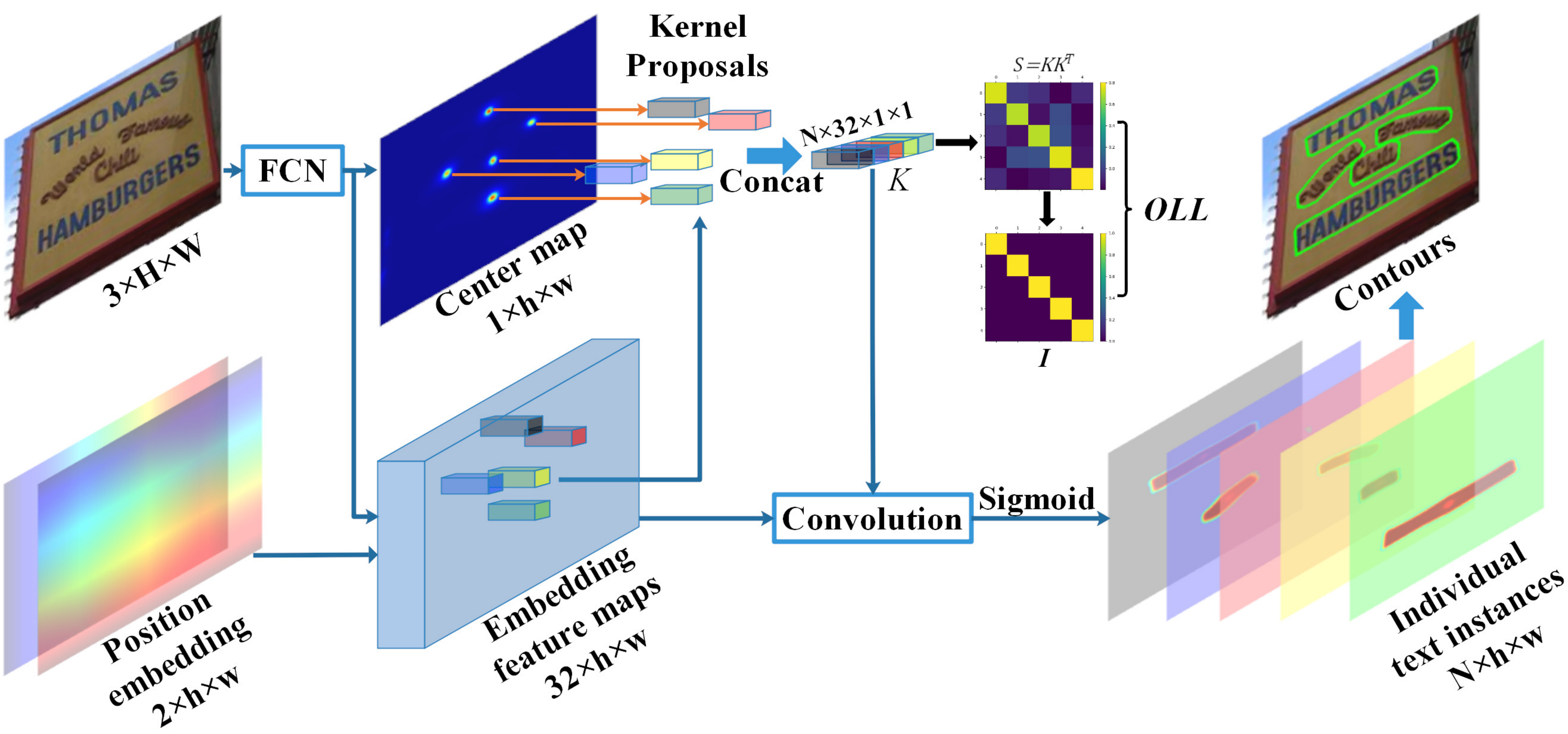}
	\end{center}%
	\caption{Framework of our method. The feature extractor Fully Convolutional Network (FCN) is detailed in Fig. \ref{fig:network}. The network will dynamically predict $N$ (\textbf{$N$ is not fixed}) key points and get the corresponding embedding vectors for generating \textbf{Kernel Proposals}. To enforce the independence between kernel proposals, we propose a novel orthogonal learning loss (OLL) via orthogonal constraints to make $ S=KK^{T} $ close to an identity matrix ($ I $). Finally, we predict text masks for generating contours. Examples of key point generation: (a) predicted heat-maps; (b) connected components (CCs) after thresholding; (c) key points that with the scores in each connected component.}
	\label{fig:pipeline}
\end{figure*}

\section{Our Method}\label{Proposed_Method}
\subsection{Overwiew}
As shown in Fig. \ref{fig:pipeline}, our method mainly consists of four components: feature extraction, kernel proposal, instance-independent feature map extraction, and contour generation. 
We utilize a fully convolutional network (FCN) \cite{FCN} combined with a feature pyramid network (FPN) \cite{FPN} to extract informative feature maps. The architecture details of our feature extraction sub-network are elaborated in Fig. \ref{fig:network}. 
To extract kernel proposals, we first get the connected component of each text in the predicted center map and select the pixel with the highest score in each component as the key point. The embedding feature maps in the corresponding position of key points are predicted kernel proposals, which will be convolved by the predicted kernel proposals for generating instance-independent feature maps. The predicted feature maps will be binarized by a pre-defined threshold to get the contours of the detected texts.

\subsection{Kernel Proposal}
For segmenting text instances in arbitrary shapes, a majority of existing methods adopted the popular FCN structure. However, they often suffer from texts with tiny intervals or unclear boundaries, as shown in Fig. \ref{fig:introduction} (a). To overcome these problems, some methods adopt embedding strategies to link \cite{TextField,PixelLink} or cluster \cite{CVPR19_Embedding,PAN} pixels into different text instances. The success of these methods lies in finding a metric function $F(*,*)$ to judge if two pixels belong to the same text, which can be formulated as

\begin{equation}
	\label{eq_embedding}
	F(e_i,e_j)
	\begin{cases} 
		1, \  T(p_i) = T(p_j), \\
		
		0, \  T(p_i) \neq T(p_j),\\
	\end{cases}
\end{equation}
where $p_i$ and $p_j$ denote the pixels in position $i$ and $j$, respectively; $T(*)$ denotes the corresponding text of the pixel; $e_i$ and $e_j$ denote the embedding features of $p_i$ and $p_j$, respectively; $F(*,*)$ is a learnable function.

From our observation, if we find an appropriate $F(*,*)$ as a binary classifier, we can classify the pixels of different text instances into instance-independent feature maps. Fortunately, it can be well implemented by a convolution layer together with a sigmoid function. Specifically, we can find a suitable convolution kernel (\ie, dynamic kernel) for one text instance, which can convolve all the embedding feature maps to classify the pixels belonging to this text instance to an instance-independent feature map. It can help us segment each text instance efficiently.

As mentioned above, we try to utilize a group of convolution kernels for classifying different text instances into instance-independent feature maps. In traditional convolution methods, the number of kernels is
fixed, making the channels of output feature maps are also fixed. However, the number of text instances often varies in different scene images. Thus, the specified number of instance-independent feature maps may induce missed or repeated detection. Inspired by dynamic Kernel methods~\cite{AdaptIS, solov2}, we propose a dynamic convolution kernels strategy (kernel proposal) to separate text instances into different instance-independent feature maps. Specifically, we firstly predict $N$ key points for $N$ text instances, then extract the feature maps according to their corresponding positions to generate kernel proposals (\ie, dynamic convolution kernels). Different from other dynamic kernel methods, our kernel proposals are orthogonal to each other, which are a set of orthogonal basis vectors. In this design, each kernel proposal contains the self-information and position information of its own text instance, not the information of other text instances. The kernel proposals will convolve the embedding feature maps to generate individual feature maps for each text instance, which is formulated as

\begin{equation}
	O = K * E=\begin{bmatrix}
		k_{0}\\...\\k_i\\...\\k_N
	\end{bmatrix}*E = \begin{bmatrix}
		p_{0}\\...\\p_i\\...\\p_N
	\end{bmatrix}\label{eq_convolve}
\end{equation}
where $O$ represents the output feature maps of, in which each channel corresponds to prediction ($ p_i $) of one text; $ k_i $ denotes the i-$ th $ kernel proposal. The convolution operation $*$ decompose the high-dimensional vector $ E $ onto the orthogonal basis $K$. $E$ denotes the embedding feature maps which contains the shared features extracted  by backbone ($ F_s $) and position embedding feature ($ F_p $) as
\begin{equation}
	E= Fbock_2(F_s \oplus F_p)
\end{equation}
where $ \oplus $ donates a concatenate operation, as shown in Fig.~\ref{fig:network}. After full training, $ F_s $ is mainly contains the features of text instances, and the non-text features tend to 0.

\begin{figure}[htbp]
	\begin{center}
		\includegraphics[width=0.98\linewidth]{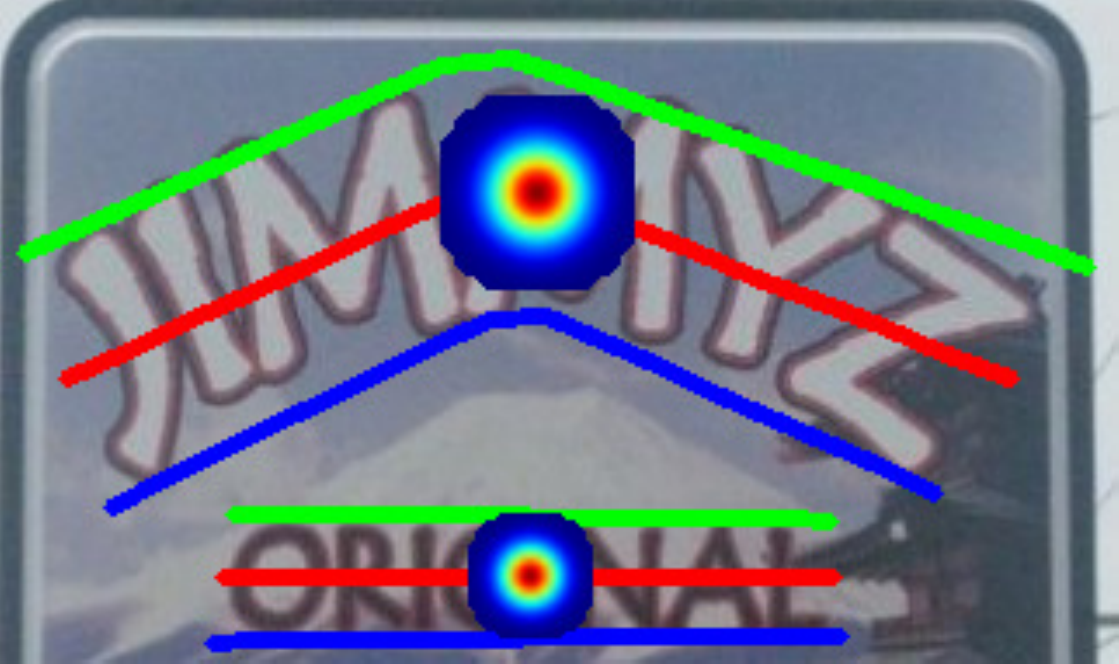}
	\end{center}
	\caption{Overview of key points labeling: top lines (green lines), bottom lines (blue lines), center lines (red lines), and center gaussian heat maps for text instances.}
	\label{fig:center_point_line}
\end{figure}

\subsection{Key Points Generation}\label{section_key_point}
If we directly utilize the function $F(*,*)$ in Eq. \ref{eq_embedding} to classify all pixel pairs, the computational complexity will be $(w \times h)^2$. To reduce the computational complexity, we only select one key point from the predicted Gaussian center maps as kernel proposals for each text instance, where one Gaussian center may correspond to one text instance. These kernel proposals contain important self-information learned by the network and key point location information by position embedding, which is the basis of our method to efficiently separate neighboring texts.

\medskip
Unlike objects in general instance segmentation, text instances do not have a closed boundary, which means that it contains a lot of regions similar to the background. It may cause our method to be sensitive to the quality of key points. To solve this problem, we adopt a broad key point for text instance. Specifically, we designate the representative center point of text instance as a center key point. As shown in Fig. \ref{fig:center_point_line}, we follow TextSnake \cite{TextSnake} to find the top line (green line) and the bottom line (blue line) of text, then calculate the centerline (red line) and extract the middle point in the centerline as the center point. Apparently, it is challenging to predict one single accurate center point. Thus we adopt the Gaussian heat-map strategy as in \cite{cornernet,center_point}, as shown in Fig. \ref{fig:center_point_line}. The radius is the minimum distance of the center point to the boundary.

\begin{figure}[htbp]
	\begin{center}
		\subfigure[Heat-map.]{
			\begin{minipage}[t]{0.31\linewidth}
				\centering
				\includegraphics[width=0.98\linewidth]{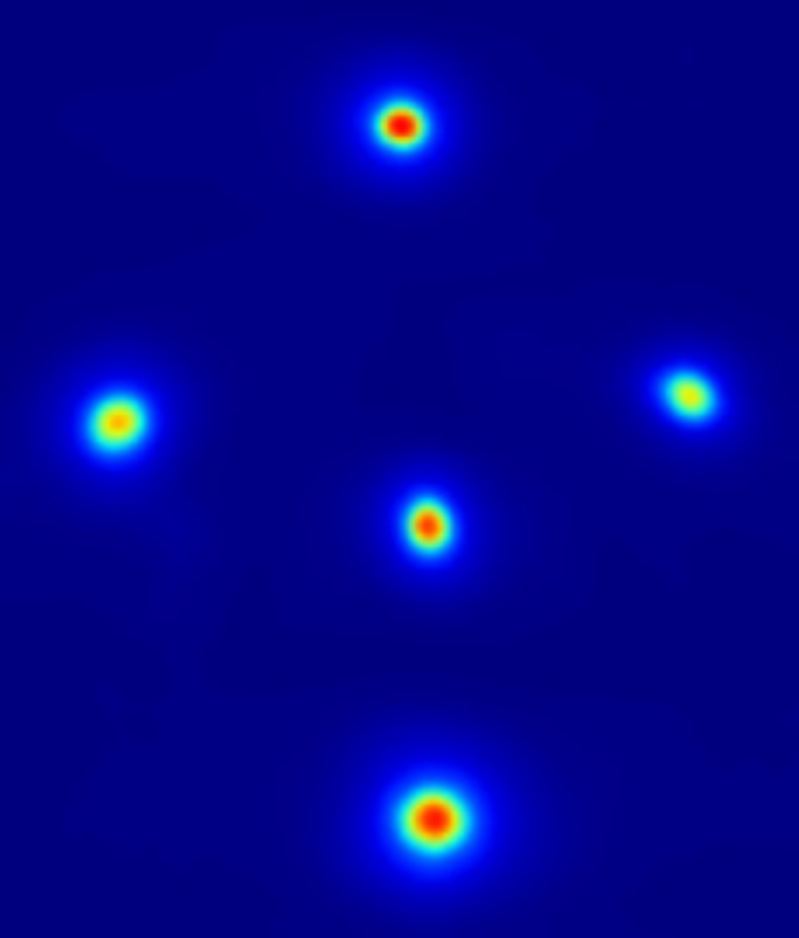}
			\end{minipage}%
		}
		\subfigure[CCs]{
			\begin{minipage}[t]{0.31\linewidth}
				\centering
				\includegraphics[width=0.98\linewidth]{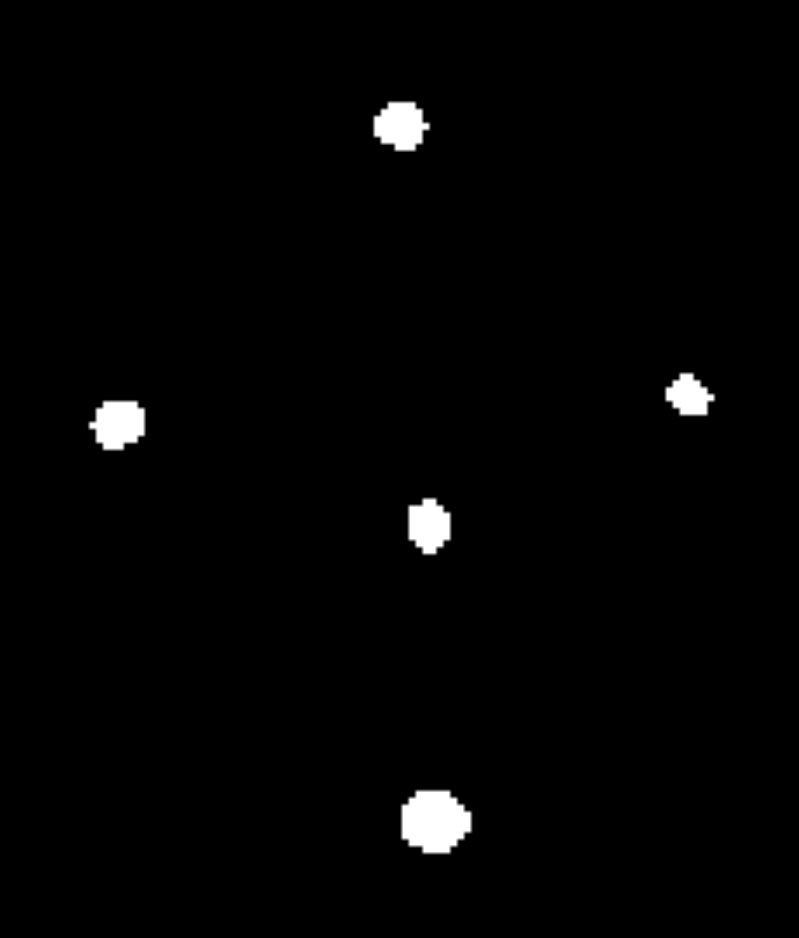}
			\end{minipage}%
		}
		\subfigure[Key points.]{
			\begin{minipage}[t]{0.31\linewidth}
				\centering
				\includegraphics[width=0.98\linewidth]{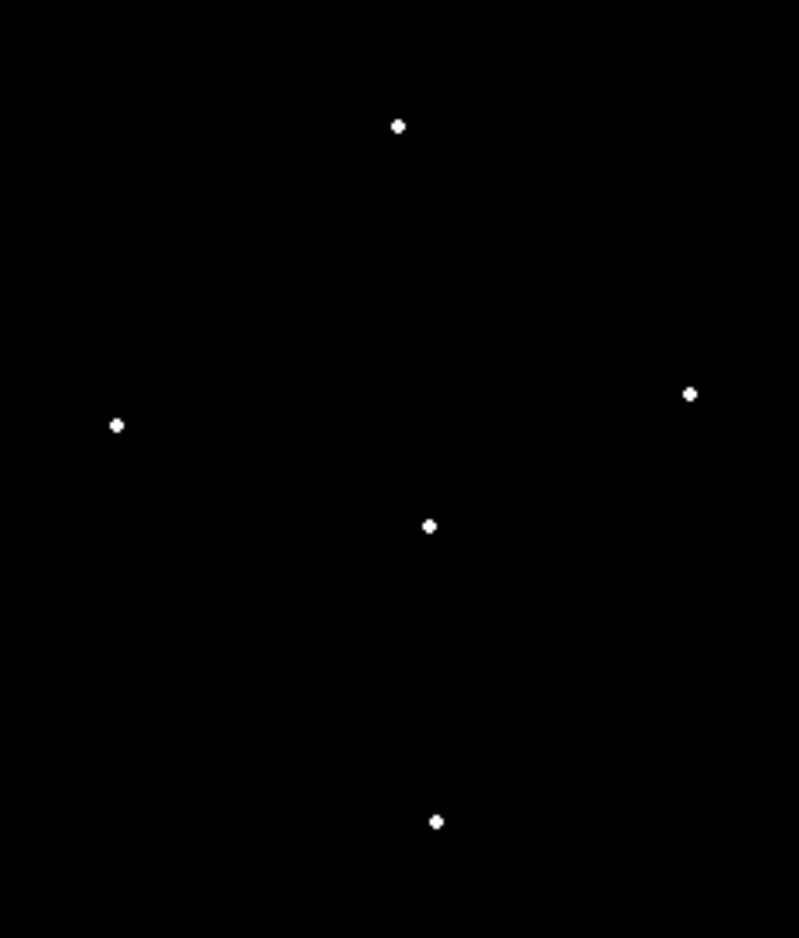}
			\end{minipage}%
		}
	\end{center}%
\caption{Examples of key point generation: (a) predicted heat-maps; (b) connected components (CCs) after thresholding; (c) key points that with the scores in each connected component.}\label{fig:get_center_point}
\end{figure}

In key point labeling, we will get a heat-map as shown in Fig. \ref{fig:get_center_point} (a). However, there still contain many redundant points for a text instance. Thus, we compute the connected component of each center via thresholding (Fig. \ref{fig:get_center_point} (b)). We will select the point with the highest score in each connected component as the final predicted key point. Kernel proposals are extracted from the embedding feature maps corresponding to the selected key points.

\subsection{Position Embedding}
Our feature extraction sub-network  
constructed on the popular FCN~\cite{FCN} (RestNet-50~\cite{ResNet} as backbone) and FPN \cite{FPN} in text detection domain. However, existing segmentation-based methods via conventional convolutional pixel embeddings can't easily distinguish identical copies of an instance \cite{semi-conv}. Inspired by \cite{semi-conv,coordconv,solov2,CVPR19_Embedding}, we also introduce a position embedding strategy in our network for keeping location information of key points, as shown in Fig. \ref{fig:pipeline}. Specifically, we adopt two channels of feature maps by embedding \textbf{x-axis} and \textbf{y-axis} position information for every pixel.

In addition, we normalize the value of position embedding to range $[-1,1]$, i.e., the pixel in x-axis 0 is set to $-1$, and the pixel in x-axis is set to 1. Mathematically, the position embedding of $x_i$ and $y_i$ can be formulated as
\begin{equation}
	x_i = \{-1+\frac{2i}{w-1} | i \in (0,w-1) \}
\end{equation}
\begin{equation}
	y_i = \{-1+\frac{2i}{h-1} | i \in (0,h-1) \}
\end{equation}
where $w$ and $h$ respectively represent the width and height of the output feature maps.

\subsection{Loss Function}
We optimize the prediction of the Gaussian center map and the segmentation of each instance with the following loss function:
\begin{equation}
	\begin{split}
		L_{c}(y, \hat{y}, \alpha) &= \alpha* L_{dice}(y, \hat{y}) +  L_{focal}(y, \hat{y}) \\&+  L_{OHEM}(y, \hat{y}) +  L_{BBCE}(y, \hat{y})
	\end{split}
\end{equation}
where $L_{dice}$ denotes Dice loss \cite{diceloss}; $L_{focal}$ denotes Focal loss \cite{focalloss}; $L_{OHEM}$ denotes Cross Entry loss with OHEM \cite{OHEM}, in which the ratio between  negative and positive pixels is 3:1; $L_{BBCE}$ denotes Balance Binary Cross Entropy loss; $ y$ denotes a prediction and $ \hat{y}  $ is its ground-truth; $ \alpha $ is used to select $L_{dice}$. For each instance,  the loss function of Gaussian center map ($ \mathcal{L}_c^{gc} $) and the loss function of the
segmentation ($ \mathcal{L}_c^{s}$) are respectively defined as
\begin{gather}
	\mathcal{L}_c^{gc} = L_{c}(y, \hat{y}, 0)\\
	\mathcal{L}_c^{s} = L_{c}(y, \hat{y}, 1)
\end{gather}

\begin{figure}[htbp]
	\subfigcapskip=0pt
	\centering
	\subfigure[Detected contour]{
		\begin{minipage}[t]{0.49\linewidth}
			\centering
			\includegraphics[width=4.25cm,height=4cm]{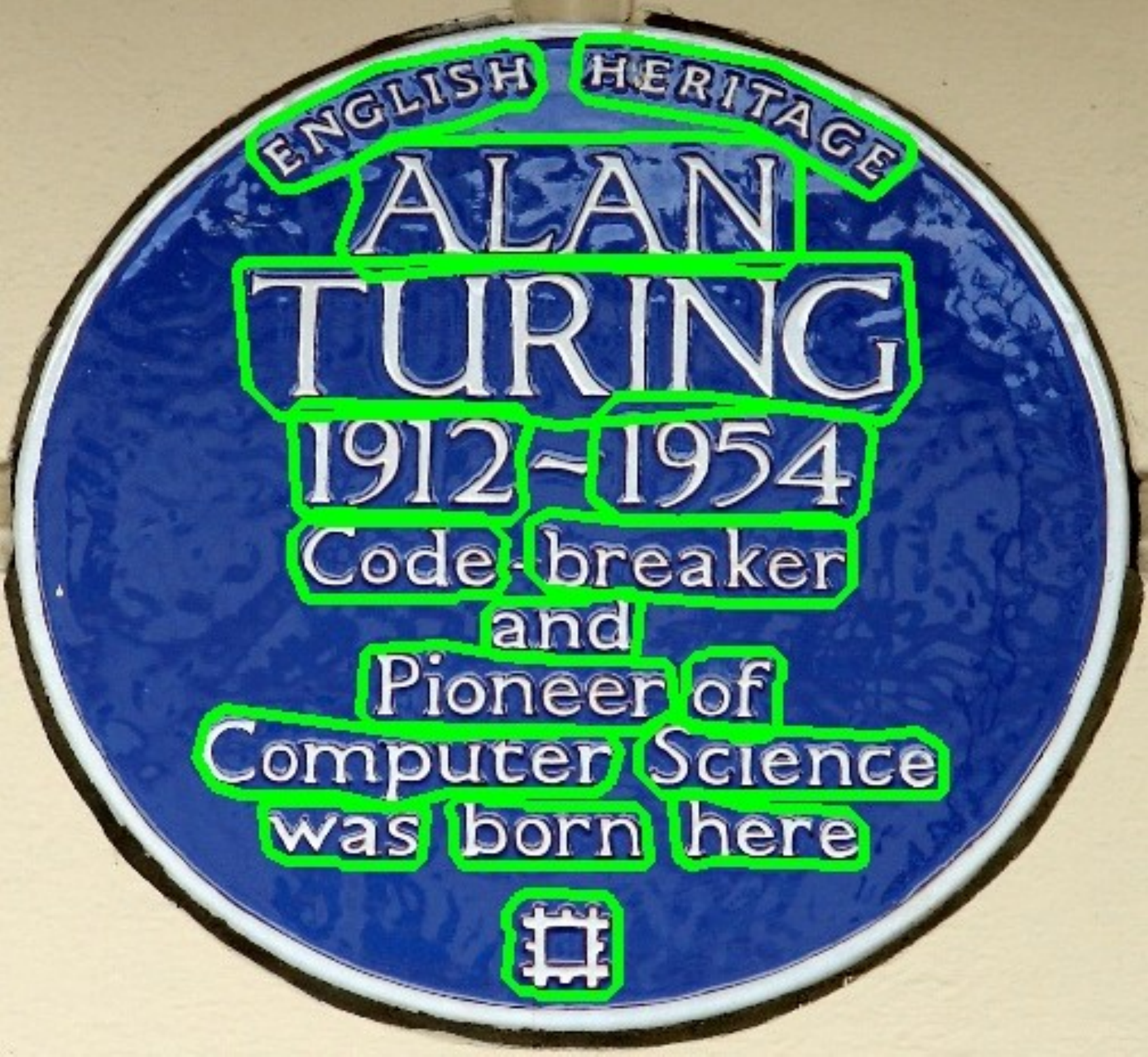}\\
		\end{minipage}%
	}%
	\subfigure[Gaussian center map]{
		\begin{minipage}[t]{0.49\linewidth}
			\centering
			\includegraphics[width=4.25cm,height=4cm]{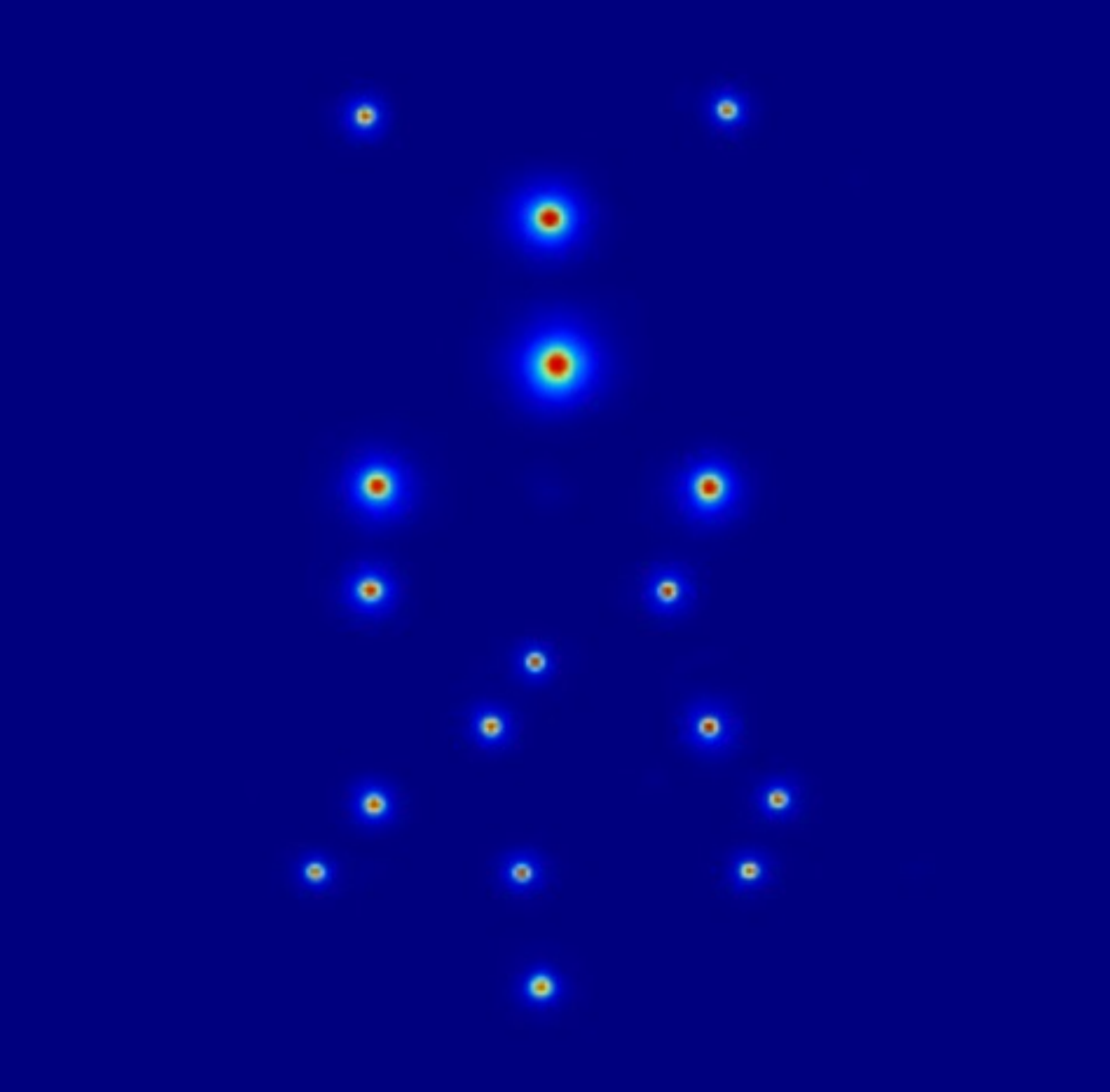}\\
		\end{minipage}
	}%
	\vspace{-3pt}
	\quad
	\subfigure[$ S $]{
		\begin{minipage}[t]{0.49\linewidth}
			\centering
			\includegraphics[width=4.25cm,height=4cm]{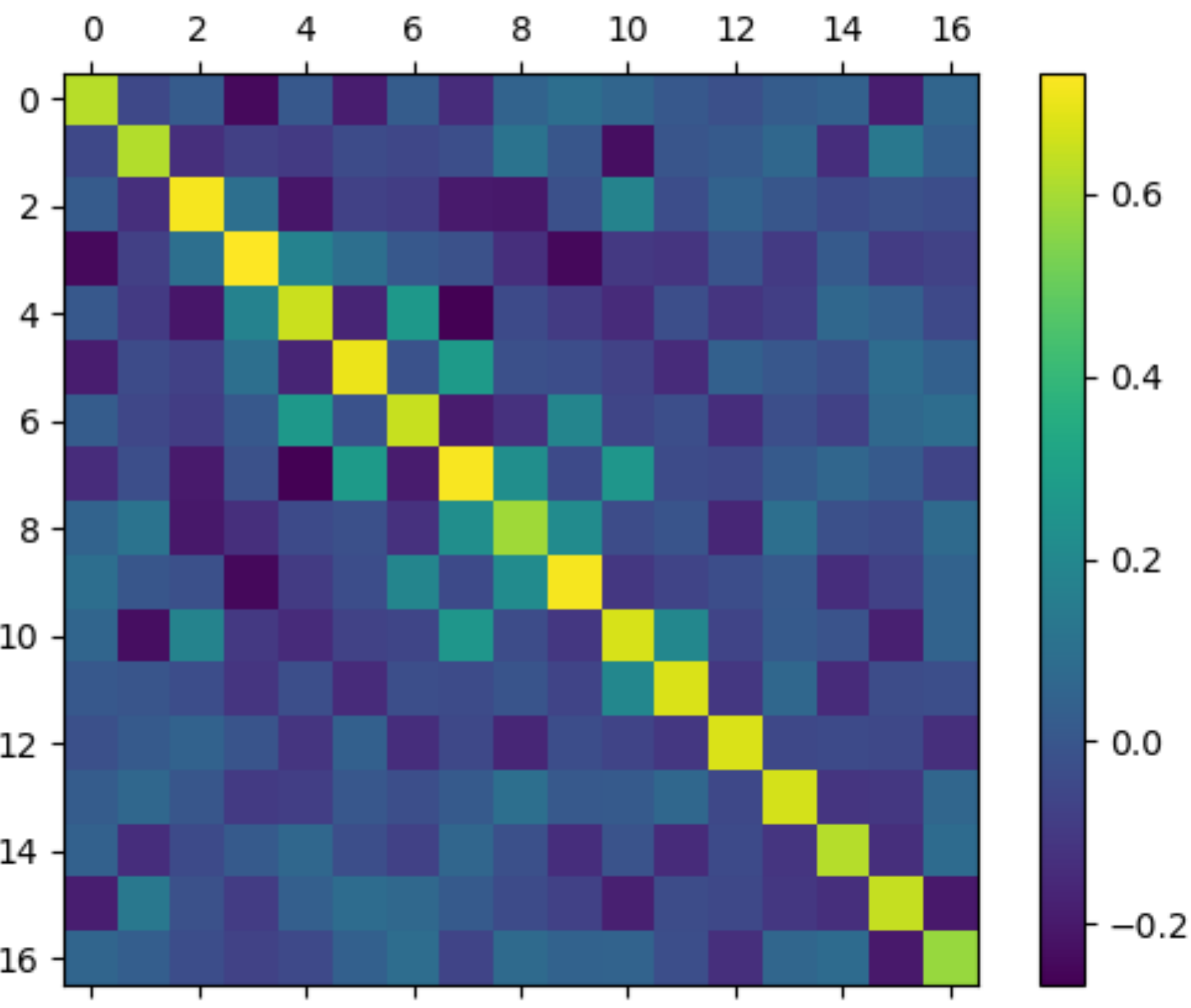}\\
		\end{minipage}
	}%
	\subfigure[$ I $]{
		\begin{minipage}[t]{0.49\linewidth}
			\centering
			\includegraphics[width=4.25cm,height=4cm]{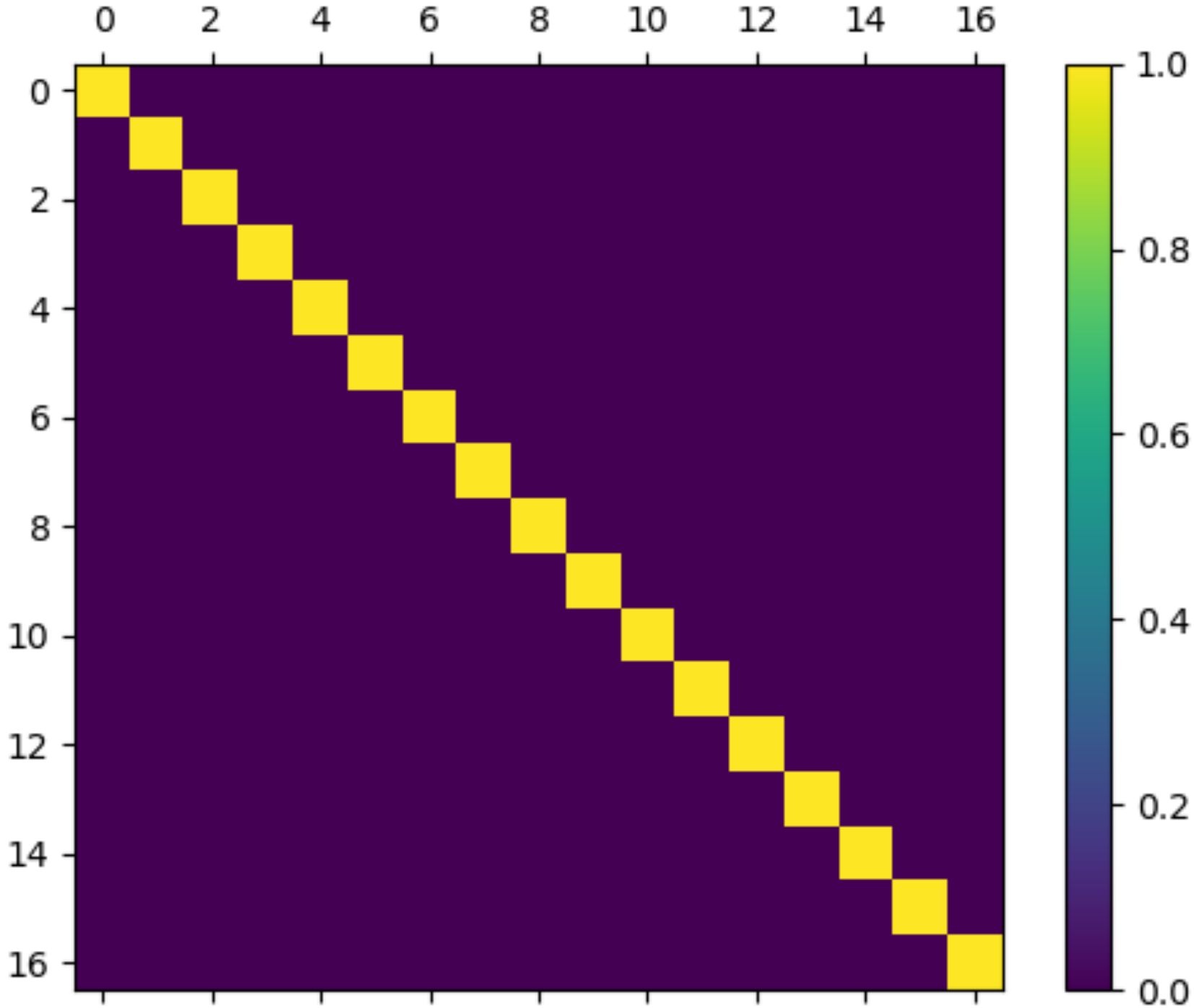}\\
		\end{minipage}
	}%
	\centering
	\caption{(a) The detected contours in the image; (b) The prediction of Gaussian center map; (c) The similarity matrix (S) between kernel proposals; (d) $ I $ is an identity matrix.}
	\label{fig:kp_loss}
\end{figure}

To separate different text instances effectively, we first need to restrict the kernel proposals to be independent of each other in one image. In this way, the kernel proposals only include the main features of their own text, not the shared features of texts. Consequently, by individually convolving all embedding feature maps with kernel proposals, our method can generate unique feature maps for individual text instances. 

In the same image, the similarity matrix between kernel proposals can be expressed as 

\begin{equation}
	S = KK ^ \mathrm{ T }=\begin{bmatrix}
		k0\\...\\k_i\\...\\k_N
	\end{bmatrix} \times \begin{bmatrix}
		k0\\...\\k_i\\...\\k_N
	\end{bmatrix}^ \mathrm{ T }
\end{equation}
where $ K $ is the set of kernel proposals in one image; $ S $ is a similarity matrix ($ N \times N $) for these kernel proposals, as shown in Fig.~\ref{fig:kp_loss}(c); $ N $ is the number of kernel proposals. To constrain the kernel proposals to be independent of each other, we propose an orthogonal learning loss ($L_{OLL}$) as
\begin{equation}
	\mathcal{L}_{OLL} = L_{dice}(S, I) +  L_{BCE}(S, I) 
\end{equation}
where $ I $ is an identity matrix, as shown in Fig.~\ref{fig:kp_loss}(d); $L_{dice}$ denotes Dice loss; $L_{BCE}$ denotes Binary Cross Entropy loss. Dice coefficient is usually used to calculate the similarity of two samples.  $L_{BCE}$ can also be used to constrain the similarity between samples to approach orthogonal relationships. 

Finally, the total loss of our method can be formulated as
\begin{equation}
	\mathcal{L} = \mathcal{L}_c^{gc} +  \mathcal{L}_c^{s} + \lambda_O* \mathcal{L}_{OLL}
\end{equation}
where $\lambda_O$ is set to 0.1.

\section{Experiments} \label{Experiments}

\begin{figure}[htbp]
	\begin{center}
		\subfigure[KPN.]{
			\begin{minipage}[t]{0.46\linewidth}
				\centering
				\includegraphics[width=1\linewidth]{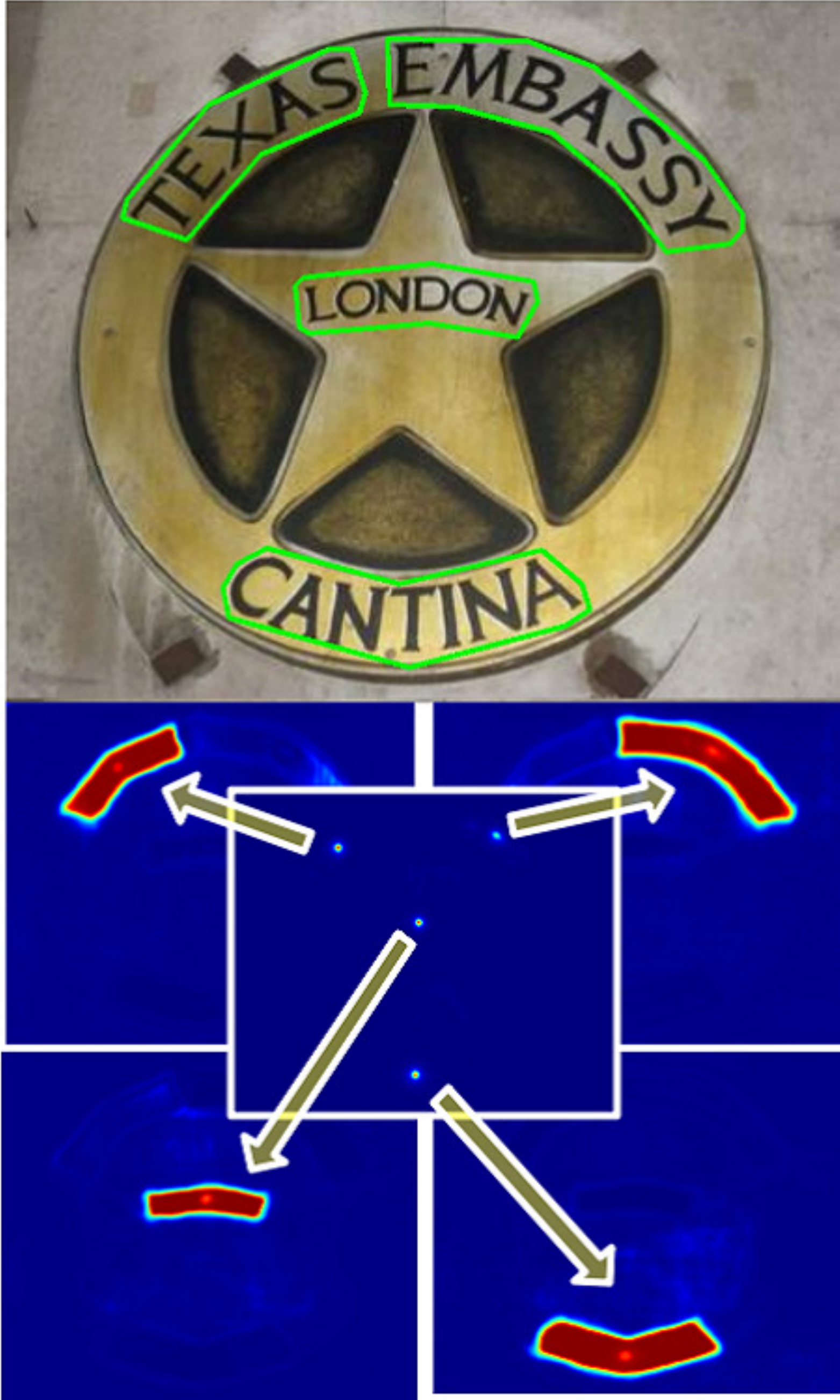}
			\end{minipage}%
		}
		\subfigure[FCN.]{
			\begin{minipage}[t]{0.46\linewidth}
				\centering
				\includegraphics[width=1\linewidth]{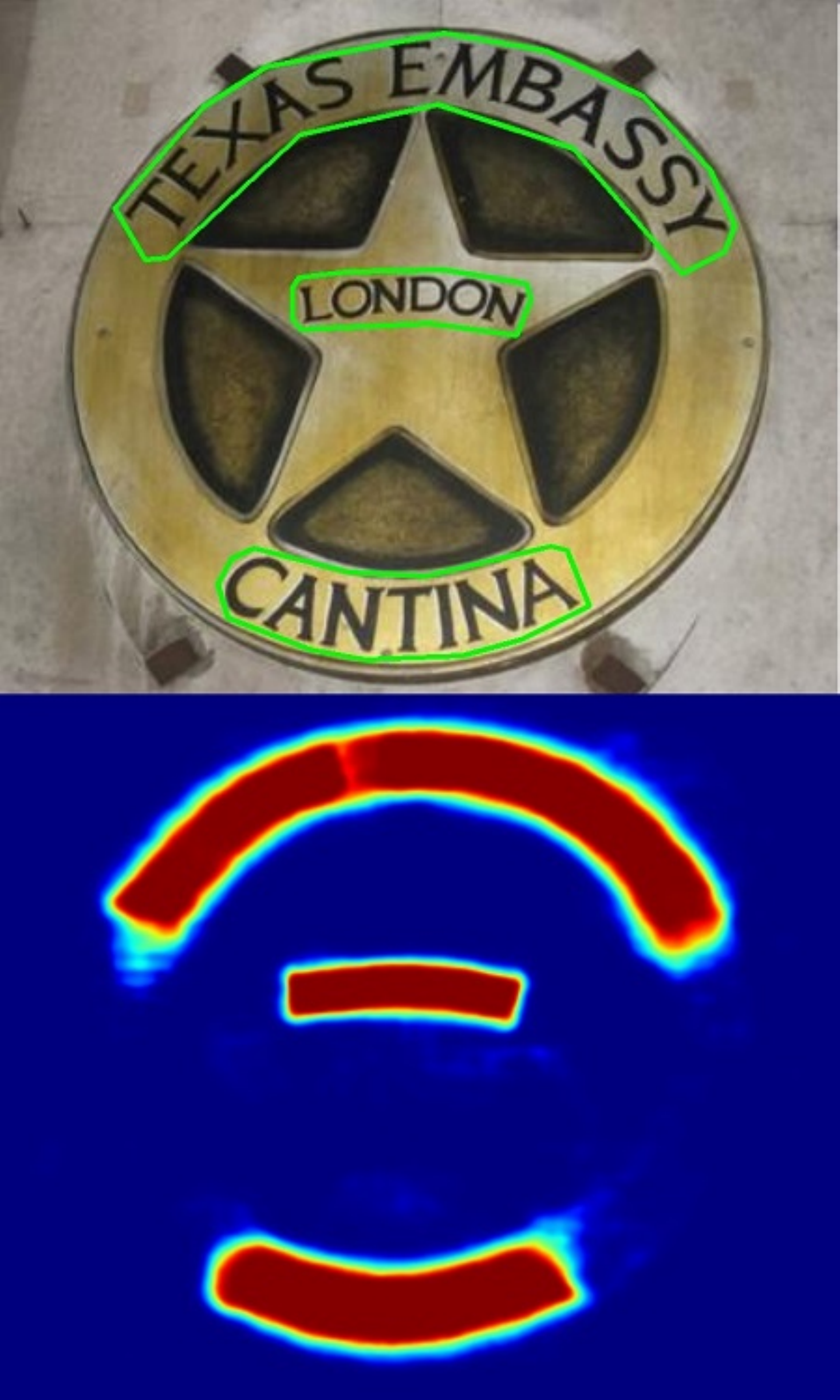}
			\end{minipage}%
		}
	\end{center}%
	\caption{Representative detecting results of KPN and FCN.}\label{fig:ablation_KPN_FCN}
\end{figure}

\subsection{Datasets}
To verify the effectiveness of the proposed KPN, we conduct experiments on three benchmark datasets: Total-Text \cite{totaltext}, CTW-1500 \cite{ctw1500}, and ICDAR 2015 \cite{IC15}. 

\medskip
\textbf{Total-Text} \cite{totaltext} consists of $1,255$ training and $300$ testing images. It is collected from various scenes, including text-like background clutter and low-contrast texts, and are word-level annotated by polygons for curve text detection task.

\medskip
\textbf{CTW-1500} \cite{ctw1500} consists of $1,000$ training and $500$ testing images. It contains both English and Chinese texts
with text-line level polygon annotations for curved text detection.

\medskip
\textbf{ICDAR 2015} \cite{IC15} consists of $1,000$ training and $500$ testing images. It is a multi-orientated and street-viewed dataset collected for the arbitrary-oriented text detection task. The annotations are word-level with four vertices.

\subsection{Implementation Details}\label{trainstep}
The pre-trained ResNet-50 \cite{ResNet} is adopted as the backbone of our network. We adopt Adam \cite{ADAM} for optimizing our method and the initial learning rate is 0.0001 and will be decayed by 0.9 per 100 epochs. Our network is firstly pre-trained on the large dataset MLT 2017 \cite{MLT} and randomly crop images into $640 \times 640$, $832 \times 832$, and $1024 \times 1024$, respectively. Then, we will fine-tune our network on the target benchmark dataset with $832 \times 832$ crop images. Following DRRG \cite{CVPR2020_DRRG}, we also adopt the general augmentation tricks, including crops, rotations, color variations, and partial flipping. 

In the training, we utilize two ways to get the key points for training kernel proposals: (1) We randomly sample one key point in each text instance. The sampling probability is computed on a Gaussian heat-map as described in Section \ref{section_key_point}; (2) We pick top-k (k=50) points in Gaussian heat-map for text instance as in Fig. \ref{fig:get_center_point}.  In testing, we select only one point of the highest score as kernel proposal for a corresponding text instance (connected component) as shown in Fig. \ref{fig:get_center_point}.
Our KPN contains two hyper-parameters: the threshold filtering center heat-map $thresh_c$, and the threshold for filtering final text instances $thresh_i$.
All experiments are performed on CPU (Intel(R) Xeon(R) E5-2620 v4 @ 2.10GHz), GPU (GTX 1080Ti 11G), and PyTorch 1.2.0.

\begin{table}[htbp]
	\renewcommand{\arraystretch}{1.3}
	\begin{center}
		\caption{Ablation study of KPN and FCN on Total-Text.}
		\label{tbablation_KPN_FCN}
		\begin{tabular}{c|cccc}
			\hline
			Method & Recall& Precision &H-means &FPS  \\
			\hline
			FCN-640 &62.65 &66.58 &64.55 &30.36 \\
			KPN-640 &\textbf{65.48} &\textbf{83.19} &\textbf{73.28} &23.15 \\
			\hline
			FCN-832 &66.56 &65.40 &65.98 &21.43 \\
			KPN-832 &\textbf{71.26} &\textbf{84.58} &\textbf{77.35} &15.17 \\
			\hline
			FCN-1024 &64.16 &67.99 &66.02 &15.27  \\
			KPN-1024 &\textbf{74.55} &\textbf{85.00} &\textbf{79.43} &10.54 \\
			\hline
		\end{tabular}
	\end{center}
\end{table}

\subsection{Ablation Study}\label{exp_ablation}
To verify our method, we conduct experiments on images with different scales. All the images will be resized into the range of [\textbf{short}, \textbf{large}]. In this section, we only train our models on Total-Text with 300 epochs and adopt Adam \cite{ADAM} as optimizer.

\subsubsection{KPN vs. FCN}
To verify the effectiveness of our KPN, we compare our KPN with an FCN-based method, which is similar to our feature extraction sub-network in Fig. \ref{fig:network}. The FCN-based method only utilizes the center branch to learn the masks of texts without the embedding branch. For fair comparisons, we train and evaluate our KPN and the FCN-based method both on Total-Text with 300 epochs. As listed in Table \ref{tbablation_KPN_FCN}, our KPN outperforms FCN on Total-Text nearly without sacrificing efficiency (FPS). The representative detection results are shown in Fig. \ref{fig:ablation_KPN_FCN}.

\begin{table}[htbp]
	\renewcommand{\arraystretch}{1.3}
	\begin{center}
		\caption{Ablation study of center region ($ C_r $), center point ($ C_p $), position embedding ($ P_{s} $) and test image scale on Total-Text (only trained on Total-Text with 300 epochs). `R', `P', and `H' represent `Recall', `Precision', and `H-mean', respectively.}
		\label{tbablation_center_region}
		\begin{tabular}{c|ccc|cccc}
			\hline
			Scale &$ C_r $&$ C_p $&$ P_e $& R& P &H &FPS  \\
			\hline
			KPN-640&{$ \checkmark $}&{$ \times $}&{$ \checkmark $}&59.36 &71.53 &64.88 &22.15\\
			KPN-640&{$ \times $}&{$ \checkmark $}&{$ \checkmark $}&\textbf{65.48} &\textbf{83.19} &\textbf{73.28} &\textbf{23.15}\\
			\hline
			KPN-832&{$ \checkmark $}&{$ \times $}&{$ \checkmark $}&66.22 &71.88 &68.93 &14.39\\
			KPN-832&{$ \times $}&{$ \checkmark $}&{$ \times $} &69.64 &82.13 &75.53 &\textbf{15.84}\\
			KPN-832&{$ \times $}&{$ \checkmark $}&{$ \checkmark $} &\textbf{71.26} &\textbf{84.58} &\textbf{77.35} &15.17\\
			\hline
			KPN-1024 &{$ \checkmark $}&{$ \times $}&{$ \checkmark $}&68.27 &71.76 &69.97 &9.76  \\
			KPN-1024&{$ \times $}&{$ \checkmark $}&{$ \times $} &73.35 &83.54 &78.11&\textbf{10.88}\\
			KPN-1024 &{$ \times $}&{$ \checkmark $}&{$ \checkmark $}&\textbf{74.55} &\textbf{85.00} &\textbf{79.43} &10.54 \\
			\hline
		\end{tabular}
	\end{center}
\end{table}

\begin{figure}[htbp]
	\subfigcapskip=5pt
	\begin{center}
		\subfigure[Center point]{
			\begin{minipage}[t]{0.95\linewidth}
				\centering
				\includegraphics[width=0.95\linewidth]{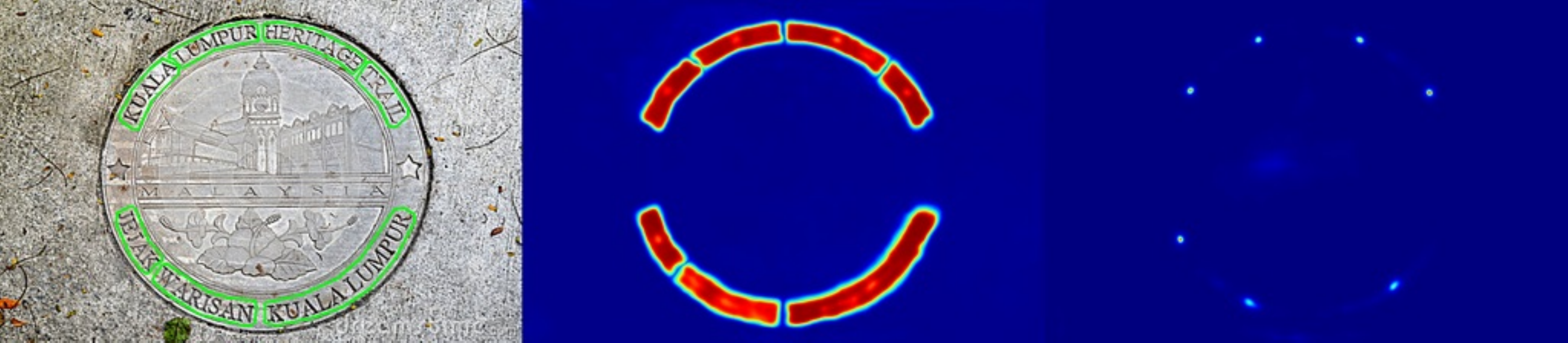}
				\centering
				\includegraphics[width=0.95\linewidth]{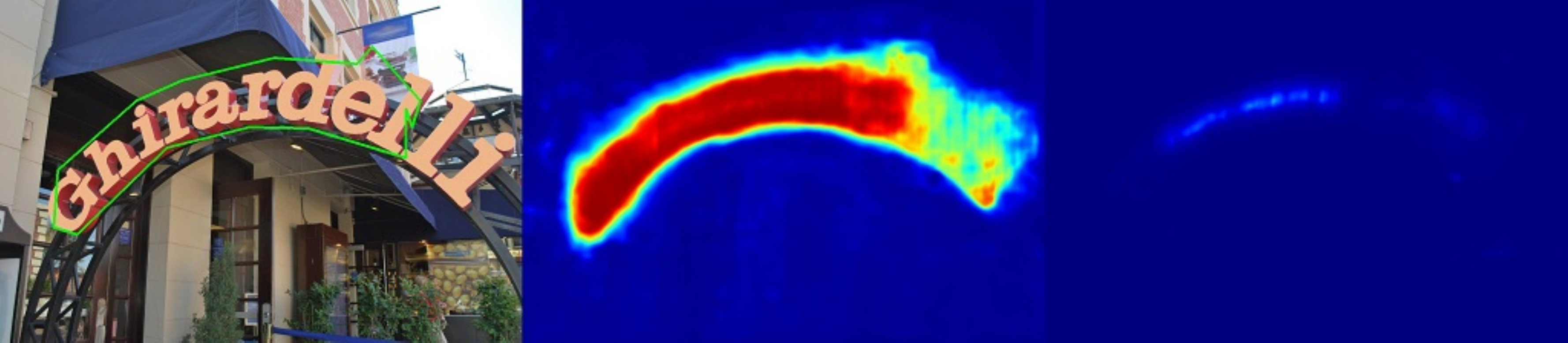}
			\end{minipage}%
		}%
		\vspace{-3pt}
		\subfigure[Center region]{
			\begin{minipage}[t]{0.95\linewidth}
				\centering
				\includegraphics[width=0.95\linewidth]{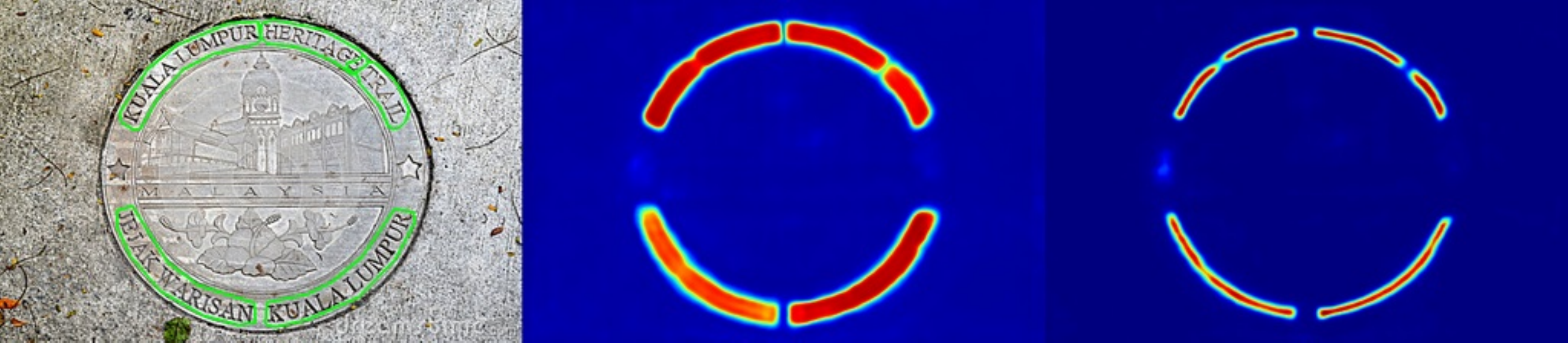}
				\centering
				\includegraphics[width=0.95\linewidth]{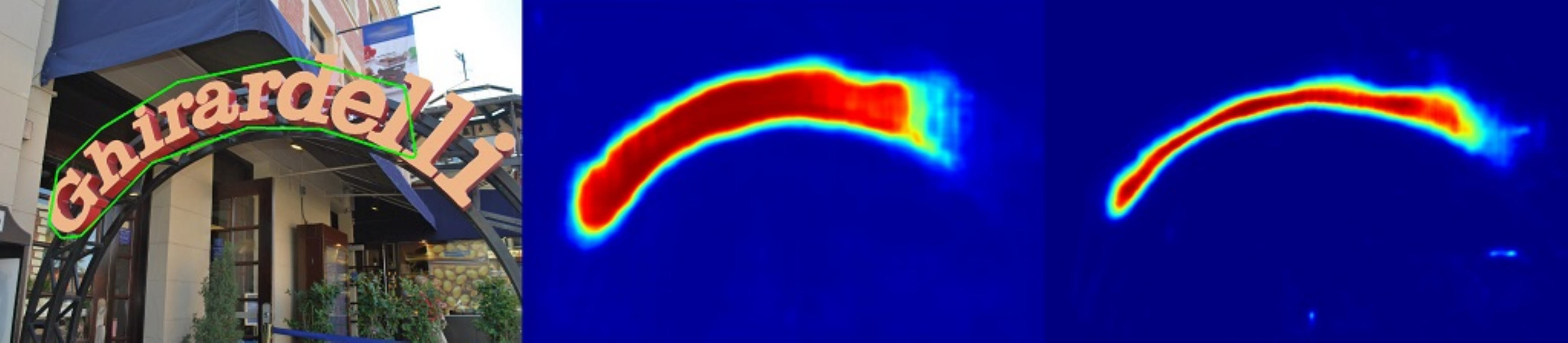}
			\end{minipage}
		}%
\end{center}%
\caption{Left column: detection results; middle column: sum of all the masks in KPN; right column: center points/regions.}
\label{fig:ablation_center_regiont}
\end{figure}

\subsubsection{Center Point vs. Center Region}
The shrinking center region of text is a popular strategy in segmentation-based methods \cite{DB, CVPR19_Embedding}. Here, we conduct an ablation study to compare our center point strategy with the center region strategy. As listed in Table~\ref{tbablation_center_region}, we can find that the center point strategy outperforms the center region strategy with a significant margin. For example, at 1024 resolution, our method outperforms the FCN by 13.41\%  in terms of H-means (KPN-1024 79.43\% vs.  FCN-1024 66.02\%). As shown in Fig.~\ref{fig:ablation_center_regiont}, we can find that our center point strategy can accurately separate adjacent text instances. But center region strategy tends to suffer from adjacent and overlapping of candidate text regions.

\subsubsection{Influence of Position Embedding}
We conduct ablation studies on Total-Text to verify the effectiveness of position embedding ($ P_{s} $). As listed in Tab.~\ref{tbablation_center_region}, without position embedding ($ P_{s} $), the performance of the proposed KPN will decrease by about 1\% to 2\% in terms of H-means. Generally speaking, location information is a fundamental basis for separating neighboring text, especially when they have similar scale and appearance. Equipped with Position Embedding, the kernel proposal and dynamic convolution kernel generated in our KPN will contain the central position information of the corresponding text instance.

\begin{figure*}[htbp]
	\subfigcapskip=0pt
	\centering
	\subfigure[Detected contour]{
		\begin{minipage}[b]{0.165\linewidth}
			\centering
			\includegraphics[width=3cm,height=6.8cm]{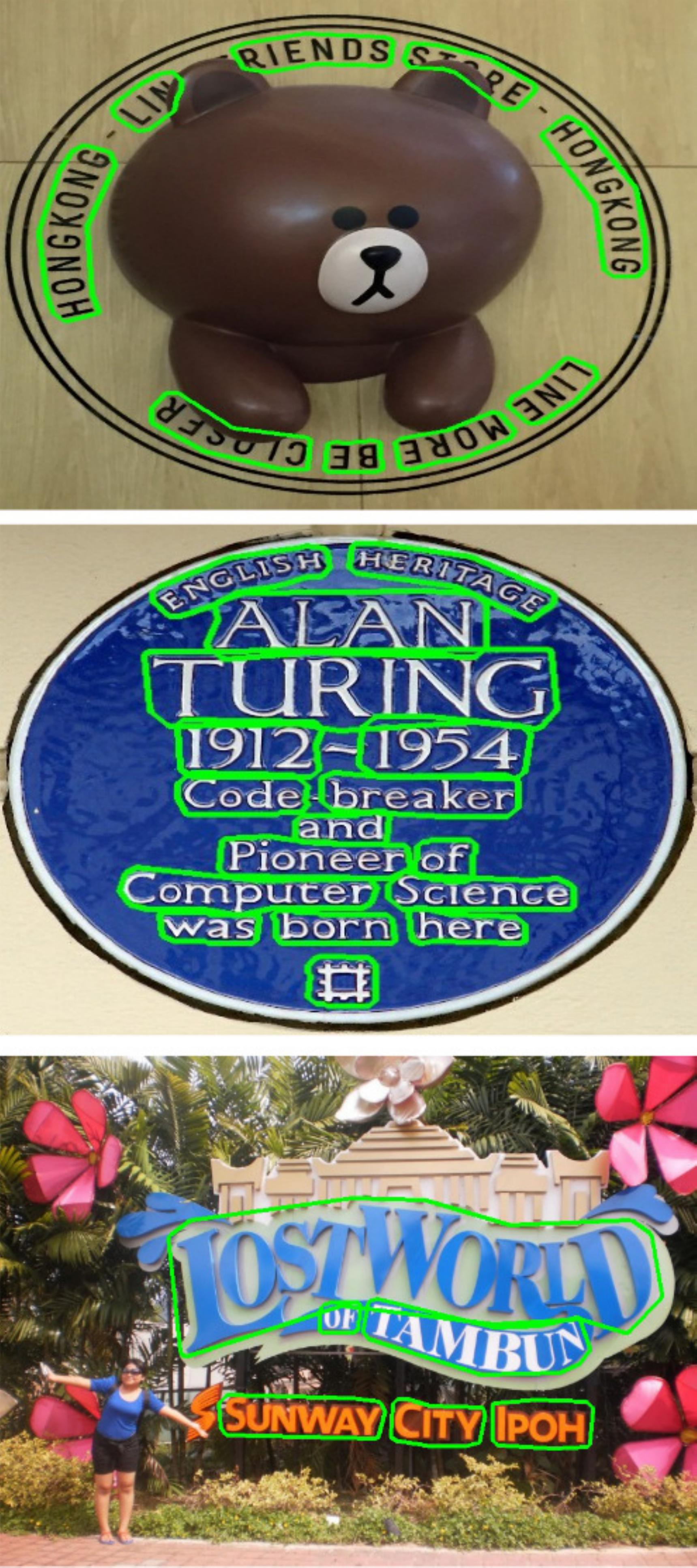}
		\end{minipage}%
	}%
	\subfigure[Gaussian center map]{
		\begin{minipage}[b]{0.165\linewidth}
			\centering
			\includegraphics[width=3cm,height=6.8cm]{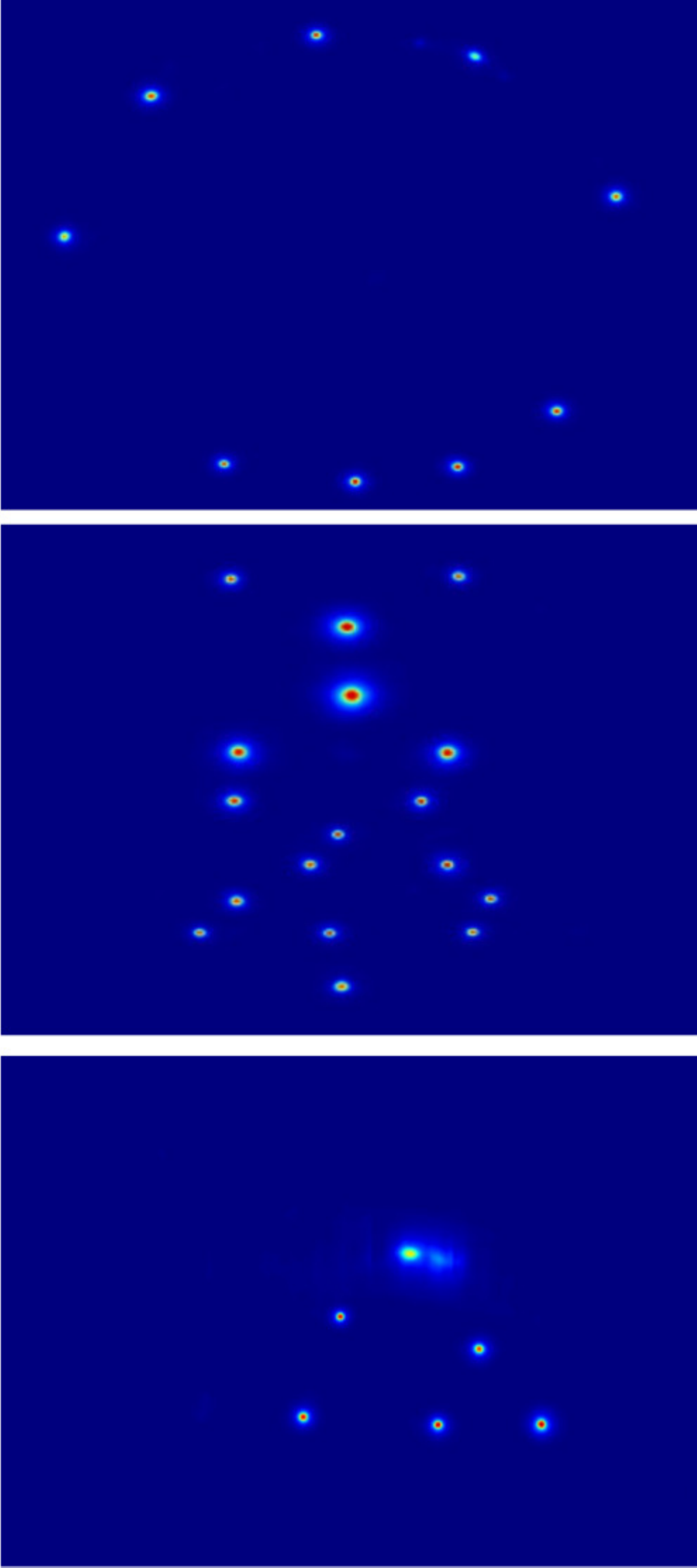}
		\end{minipage}%
	}%
	\subfigure[$ S $]{
		\begin{minipage}[b]{0.165\linewidth}
			\centering
			\includegraphics[width=3.0cm,height=6.8cm]{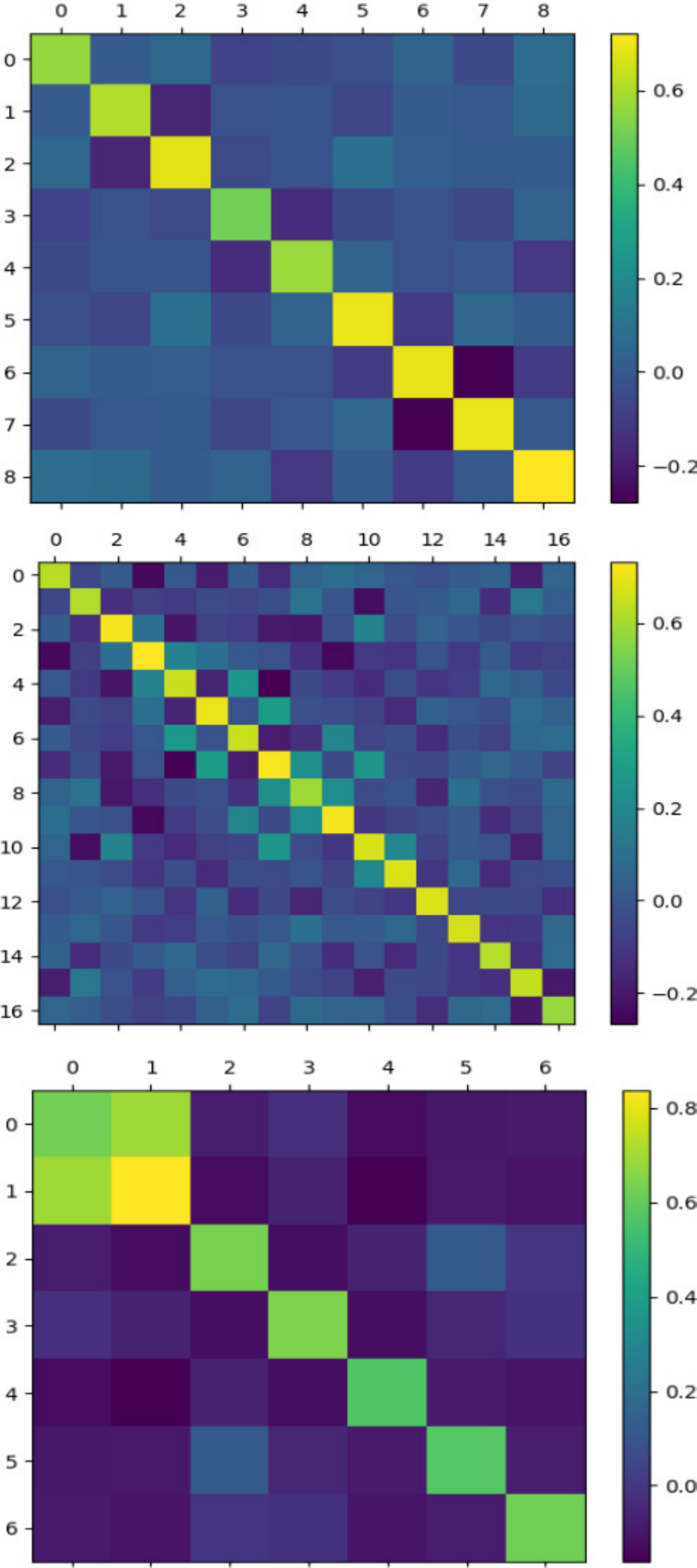}
		\end{minipage}%
	}%
	\subfigure[Detected contour$ ^{*} $]{
		\begin{minipage}[b]{0.165\linewidth}
			\centering
			\includegraphics[width=3cm,height=6.8cm]{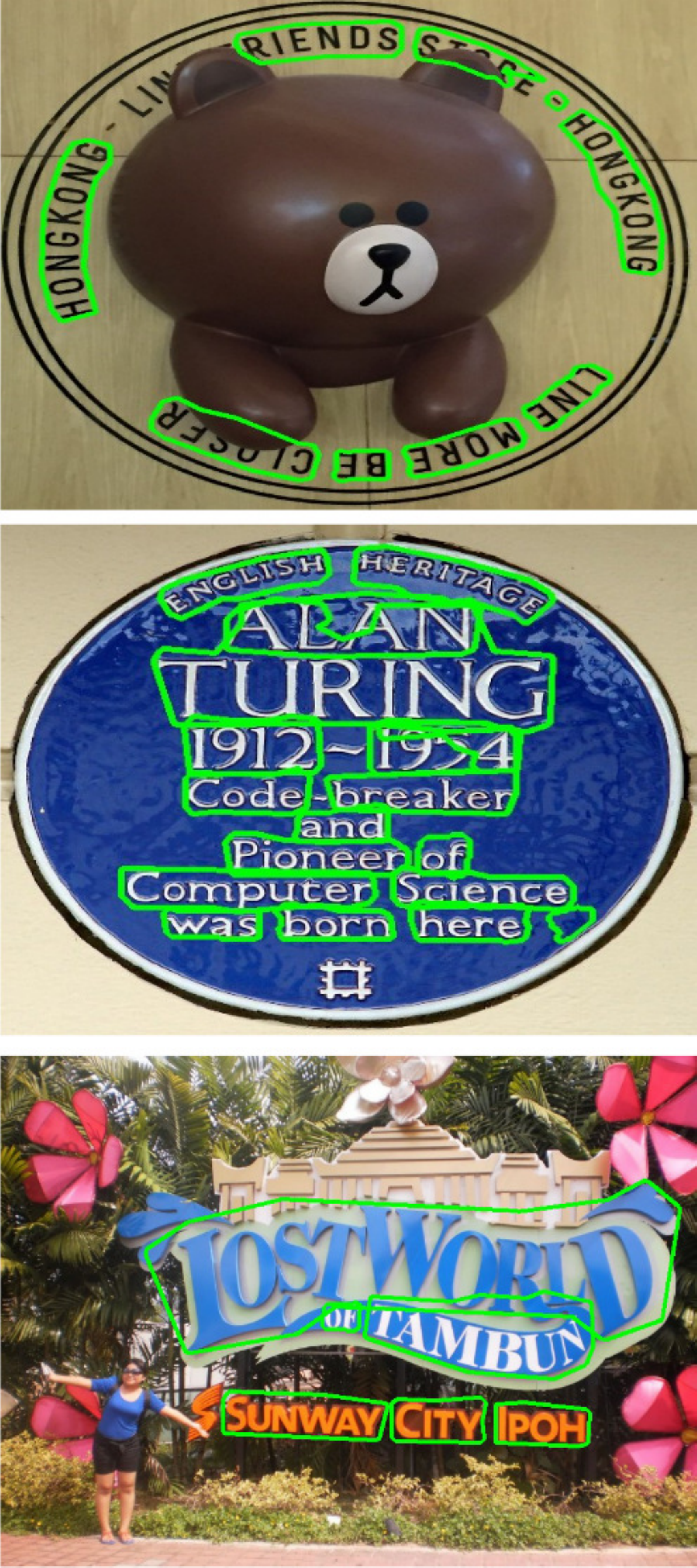}
		\end{minipage}%
	}%
	\subfigure[Gaussian center map$ ^{*} $]{
		\begin{minipage}[b]{0.165\linewidth}
			\centering
			\includegraphics[width=3cm,height=6.8cm]{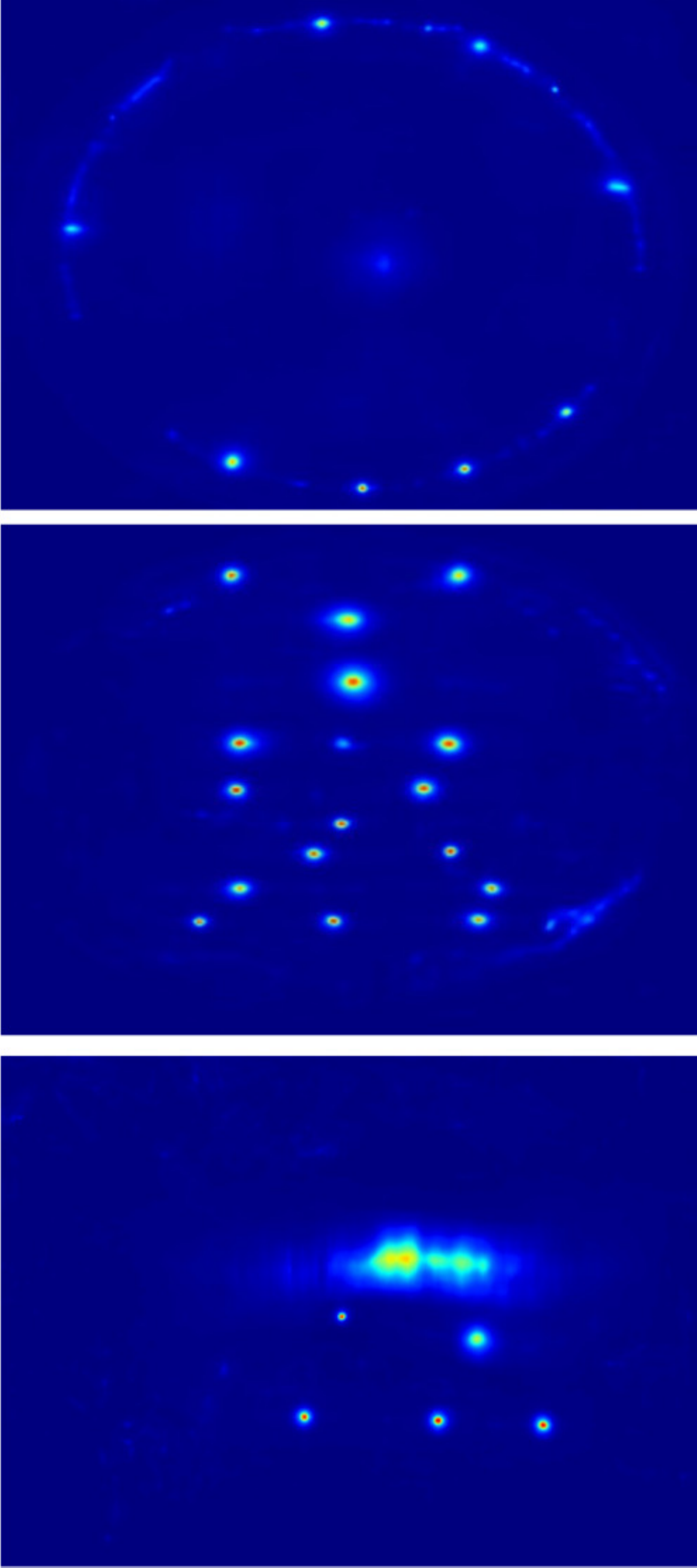}
		\end{minipage}%
	}%
	\subfigure[$ S^{*} $]{
		\begin{minipage}[b]{0.165\linewidth}
			\centering
			\includegraphics[width=3cm,height=6.8cm]{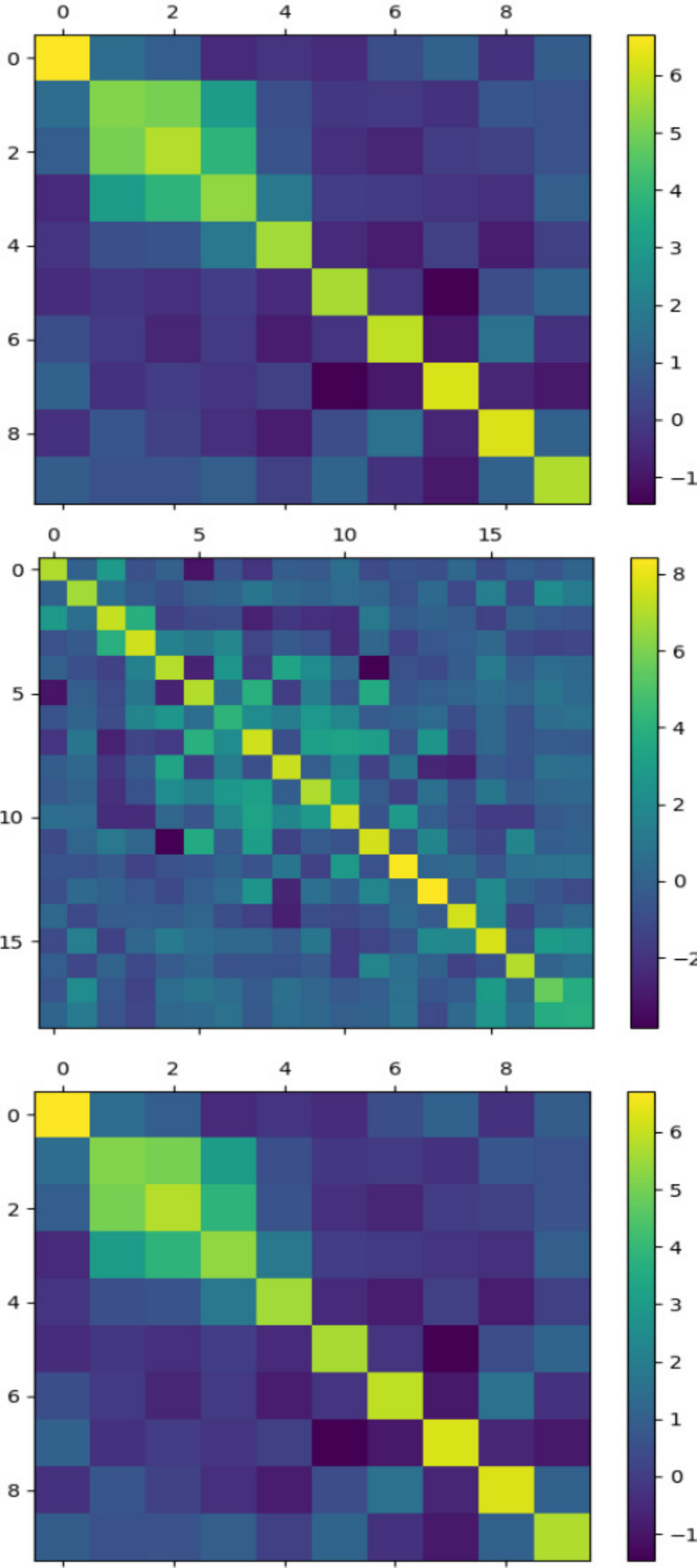}
		\end{minipage}%
	}%
	\centering
	\caption{  Representative visual results on Total-Text dataset with or without orthogonal learning loss (OLL). $ ^{*} $ means OLL is not adopted in training.}
	\label{fig:kps_exp}
\end{figure*}

\begin{table}[htbp]
	\renewcommand{\arraystretch}{1.3}
	\begin{center}
		\caption{Ablation study of orthogonal learning loss (OLL)}
		\label{tb:ablation_kploss}
		\begin{tabular}{c|c|cccc}
			\hline
			Scale &$ OLL $& Recall& Precision &H-means \\
			\hline
			KPN-640&{$ \times $}&\textbf{67.86} &73.97 &70.78\\
			KPN-640&{$ \checkmark $}&65.48 &\textbf{83.19} &\textbf{73.28} \\
			\hline
			KPN-832&{$ \times $}&\textbf{71.62} &75.20 &73.36\\
			KPN-832&{$ \checkmark $} &71.26 &\textbf{84.58} &\textbf{77.35} \\
			\hline
		\end{tabular}
	\end{center}
	\vspace{-1.0em}
\end{table}

\subsubsection{Influence of orthogonal learning loss (OLL)}
We conduct ablation studies on Total-Text to verify the influence of orthogonal learning loss (OLL). As listed in Tab.~\ref{tb:ablation_kploss}, the OLL greatly improve the detection performance (by 2.5\% in KPN-640 and 3.99\% in KPN-832 in terms of H-means). In Fig.~\ref{fig:kps_exp}, we also show some visualized results of similarity matrix ($ S $) with or without orthogonal learning loss in model training. From Fig.~\ref{fig:kps_exp}, we can find that similarity matrix ($ S $) approaches an identity matrix ($ I $) with orthogonal learning loss for training, and the detection contours are more accurate. In contrast, the false detection and overlap detection apparently increase without the help of OLL, as shown in the second row of  Fig.~\ref{fig:kps_exp}(d). This is why the Recall (R) of detection is high without OLL, as listed in Tab.~\ref{tb:ablation_kploss}. In a few cases, there will be one text with multiple kernel proposals. As shown in the third row of Fig.~\ref{fig:kps_exp} (b) and (e), there are two kernel proposals for text instance "LOSTWORLD" because two centers are predicted for it. In this case, these two kernel proposals will have high similarity, as shown in the third row of Fig.~\ref{fig:kps_exp} (c) and (f). So, the same text instance will be predicted if we use these two kernel proposals to convolve embedding feature maps. With the independence restriction via orthogonal constraints, the kernel proposals can only contain important self-information and position information of its text, guaranteeing the effectiveness of classifying different texts into instance-independent maps.

\subsection{Comparisons with the State-of-the-arts}
Comparisons of the proposed KPN with the state-of-the-art methods (SOTAs) on Total-Text \cite{totaltext}, CTW-1500 \cite{ctw1500}, and ICDAR 2015 \cite{IC15} are listed in Table \ref{table:TotalText}, \ref{table:CTW1500}, \ref{table:tbICDAR15}. Total-Text and CTW-1500 consists of images with curve texts, ICDAR 2015 consists of images with quadrilateral texts. Notably, in Table \ref{table:TotalText}, \ref{table:CTW1500}, \ref{table:tbICDAR15}, `Ext' denotes extra training data for pretraining, `Syn' represents SynText dataset for pretraining, `MLT' represents ICDAR2017-MLT dataset for pretraining, and `MLT+' represents ICDAR 2017-MLT dataset and additional datasets for pretraining.

\begin{table}[htbp]
	\begin{center}
		\renewcommand{\arraystretch}{1.2}
		\caption{Experimental results on Total-Text. $^*$ and `MS' represents multi-scale test and $ \dagger $ denotes the TextSpotter-based methods.}
		\label{table:TotalText}%
		\begin{tabular}{m{2.1cm}<{\centering}
				|m{0.99cm}<{\centering}
				|m{0.55cm}<{\centering}
				|m{0.5cm}<{\centering}
				|m{0.5cm}<{\centering}
				|m{0.5cm}<{\centering}
				|m{0.5cm}<{\centering}}
			\hline
			Methods &Paper&Ext&R & P& H&FPS \\
			\hline
			TextSnake \cite{TextSnake} &ECCV'18&Syn&74.5 &82.7 & 78.4& -\\
			FTSN \cite{FTSN} &ICPR'18&Syn &78.0 &84.7 & 81.3 &- \\
			MSR \cite{MSR}  &IJCAI'19&Syn&73.0  &85.2 & 78.6& 4.3\\
			TextField \cite{TextField}&TIP'19&Syn  &79.9 &81.2 & 80.6& 6\\
			SegLink++ \cite{SegLink++} &PR'19&Syn &80.9 &82.1 & 81.5 & \\
			ATTR \cite{CVPR19_ATRR}&CVPR'19&-  &76.2  &80.9 & 78.5& -\\
			CSE \cite{CVPR19_CSE} &CVPR'19&MLT &79.1 &81.4 & 80.2& 0.42\\
			PSENet-1s \cite{CVPR19_PSENet} &CVPR'19&MLT  &77.96 &84.02 & 80.87&3.9 \\
			LOMO \cite{CVPR19_LOMO} &CVPR'19 &MLT+ &75.7 &88.66 & 81.6& 4.4 \\
			LOMO* \cite{CVPR19_LOMO}&CVPR'19 &MLT+&79.3 &87.6 & 83.3& \\
			CRAFT \cite{CRAFT}&CVPR'19&Syn  &79.9 &87.6 & 83.6& -\\
			TextDragon$ \dagger $ \cite{TextDragon}  &ICCV'19&MLT+&75.7 &85.6 & 80.3&- \\
			PAN-640 \cite{PAN} &ICCV'19&Syn &81.0 &\textbf{89.3} &85.0& \textbf{39.6}\\
			DB \cite{DB} &AAAI'20&Syn &82.5 &87.1 &84.7 &32\\
			ContourNet \cite{CVPR2020_ContourNet}&CVPR'20&-  &83.9 &86.9 &85.4& 3.8\\
			DRRG \cite{CVPR2020_DRRG} &CVPR'20&MLT &84.93 &86.54 &85.73 & -\\
			ABCNet$ \dagger $ \cite{ABCNet} &CVPR'20&MLT+ &81.3 &87.9 &84.5 & -\\
			TextRay\cite{TextRay}&MM'20&- &77.9 &83.5 &80.06 & -\\
			\hline
			
			\hline
			\textbf{KPN-640} &-&MLT&82.33 &88.00 &85.07 &22.73 \\
			\textbf{KPN-832} &-&MLT&85.6 &88.66 &87.11 &15.03 \\
			\textbf{KPN MS} &-&MLT&\textbf{87.04} &88.17 &\textbf{87.60} &- \\
			\hline
		\end{tabular}
	\end{center}%
\end{table}
\begin{table}[htbp]
	\begin{center}
		\renewcommand{\arraystretch}{1.2}
		\caption{Experimental results on  CTW-1500.}
		\label{table:CTW1500}
		\begin{tabular}{m{1.9cm}<{\centering}
				|m{0.99cm}<{\centering}
				|m{0.6cm}<{\centering}
				|m{0.5cm}<{\centering}
				|m{0.5cm}<{\centering}
				|m{0.5cm}<{\centering}
				|m{0.5cm}<{\centering}}
			\hline
			Methods &Paper&Ext&R & P& H &FPS \\
			\hline
			TextSnake \cite{TextSnake} &ECCV'18&Syn&85.3 &67.9 & 75.6 & -\\%
			CTD \cite{CTD} &PR'19&Syn&65.2 &74.3 & 69.5 &-\\
			SegLink++ \cite{SegLink++} &PR'19&Syn&79.8 &82.8 & 81.3&- \\
			TextField$^*$ \cite{TextField} &TIP'19&Syn&79.8 &83.0 & 81.4& 6\\
			MSR\cite{MSR}  &IJCAI'19&Syn&79.0 &84.1 & 81.5 & 4.3\\
			CSE \cite{CVPR19_CSE}  &CVPR'19&MLT&76.0 &81.1& 78.4& 0.38\\
			LOMO$^*$ \cite{CVPR19_LOMO} &CVPR'19&MLT+ &89.2 &69.6 & 78.4& 4.4\\
			LSAE \cite{CVPR19_Embedding} &CVPR'19&Syn&77.8 &82.7 & 80.1 & 3\\
			ATRR \cite{CVPR19_ATRR}  &CVPR'19&-&80.2 &80.1 & 80.1 & -\\
			PSENet-1s \cite{CVPR19_PSENet}  &CVPR'19&MLT&79.7 &84.8 & 82.2& 3.9\\
			CRAFT \cite{CRAFT}  &CVPR'19&Syn &81.1 &86.0 &83.5& -\\
			TextDragon$ \dagger $\cite{TextDragon} &ICCV'19&MLT+ &82.8 &84.5 &83.6& -\\
			PAN-640 \cite{PAN}  &ICCV'19&Syn &81.2 &86.4 &83.7& \textbf{39.8}\\
			DB \cite{DB}  &AAAI'20&Syn&80.2 &\textbf{86.9} &83.4 &22\\
			ContourNet \cite{CVPR2020_ContourNet} &CVPR'20&- &84.1 &83.7 &83.9& 4.5\\
			ABCNet$ \dagger $ \cite{ABCNet} &CVPR'20&MLT+ &83.4 &84.4 &81.4 & -\\
			DRRG \cite{CVPR2020_DRRG} &CVPR'20&MLT &83.02 &85.93 &84.45 & -\\
			\hline
			
			\hline
			\textbf{KPN-640} &-&MLT&82.86 &84.03 &83.44 &24.25 \\
			\textbf{KPN-832} &-&MLT&84.19 &84.36 &84.27 &16.30\\
			\textbf{KPN MS} &-&MLT&\textbf{86.44} &84.04 &\textbf{85.22} &-\\
			\hline
		\end{tabular}
	\end{center}%
\end{table}

\begin{figure*}[htbp]
	\begin{center}
		\subfigure[Total-Text.]{
			\begin{minipage}[t]{0.31\linewidth}
				\centering
				\includegraphics[width=0.98\linewidth]{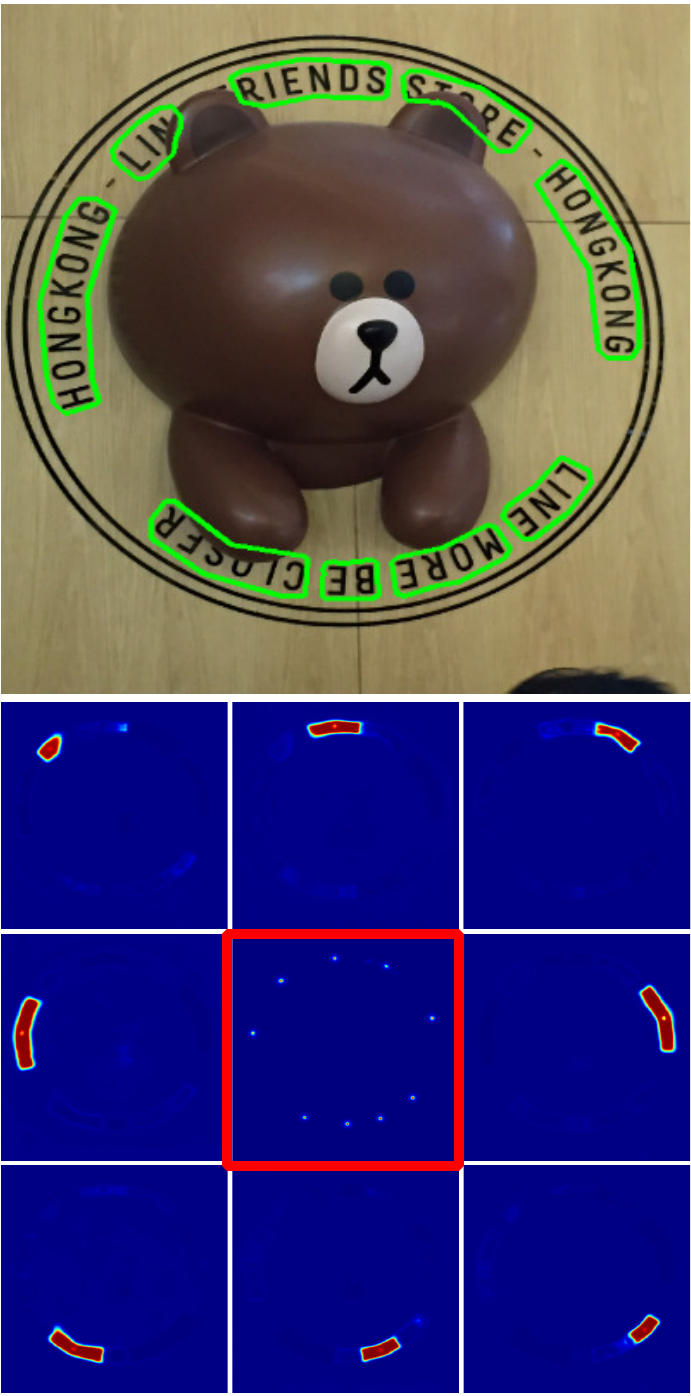}
			\end{minipage}%
		}
		\subfigure[CTW-1500.]{
			\begin{minipage}[t]{0.31\linewidth}
				\centering
				\includegraphics[width=0.98\linewidth]{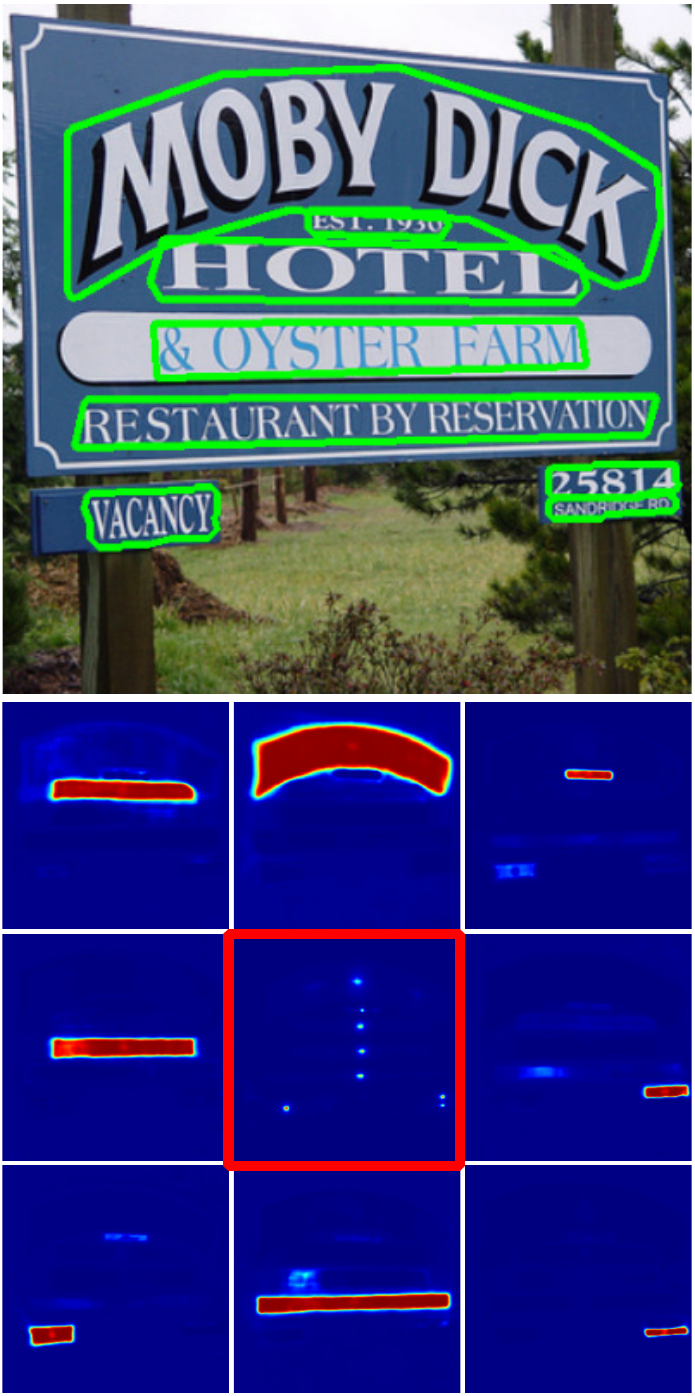}
			\end{minipage}%
		}
		\subfigure[ICDAR 2015.]{
			\begin{minipage}[t]{0.31\linewidth}
				\centering
				\includegraphics[width=0.98\linewidth]{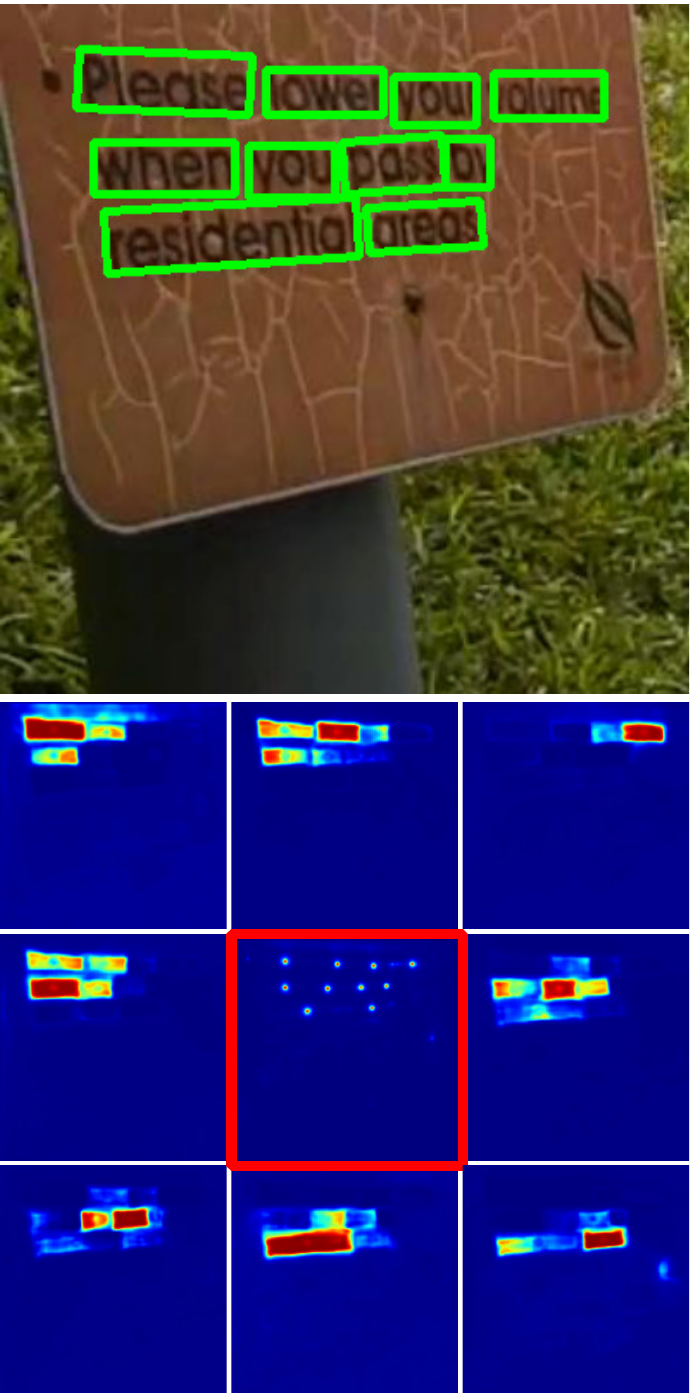}
			\end{minipage}%
		}
	\end{center}%
	\vspace{-1.5em}
	\caption{The detected results on Total-Text, CTW-1500, and ICDAR 2015. The second row shows the predicted center point maps (in center red boxes), and the individual text instances in different channels.}
	\label{fig:experiment}
	\vspace{-1.2em}
\end{figure*}

\begin{table}[htbp]
	\begin{center}
		\renewcommand{\arraystretch}{1.3}
		\caption{Experimental results on ICDAR 2015.}
		\label{table:tbICDAR15}
		\begin{tabular}{m{2.0cm}<{\centering}
				|m{0.99cm}<{\centering}
				|m{0.6cm}<{\centering}
				|m{0.5cm}<{\centering}
				|m{0.5cm}<{\centering}
				|m{0.5cm}<{\centering}
				|m{0.5cm}<{\centering}}
			\hline
			Methods &Paper&Ext&R & P& H &FPS \\
			\hline
			DDR$^*$ \cite{DDR} &ICCV'17&-&80.0 &88.0 &83.8 &1.1\\
			MCN \cite{MCN}&CVPR'18&Syn &80 &72 &76&-\\
			RRPN$^*$ \cite{RRPN} &TMM'18&-&77 &84 &80 &3.3\\
			TextSnake \cite{TextSnake} &ECCV'18&Syn&84.90 &80.40 &82.6 &1.1\\
			Textboxes++$^* \dagger$\cite{textboxes++} &TIP'18&Syn&78.50 &87.80 &82.90 &2.3\\
			PixelLink \cite{PixelLink} &AAAI'18&Syn&82.0 &85.5 &83.70 &3.\\
			FTSN \cite{FTSN}  &ICPR'18&Syn&80.0 &88.6 &84.1 &2.5\\
			IncepText \cite{IncepText} &IJCAI'18&-&80.6 &90.5 &85.3\\
			FOTS$ \dagger $ \cite{FOTS} &CVPR'18&MLT+&82.04 &88.84 &85.31 &7.8\\
			SegLink++ \cite{SegLink++} &PR'19&Syn&80.3 &83.7 & 82.0&7.1 \\
			TextField$^*$ \cite{TextField}   &TIP'19&Syn&83.9 &84.3 &84.1 &1.8\\
			PAN \cite{PAN}  &ICCV'19&Syn&81.9 &84.0 &82.9& \textbf{26.1} \\
			TextDragon$ \dagger $\cite{TextDragon}  &ICCV'19&MLT+&84.82 &81.82 &83.05 &-\\
			PSENet-1s \cite{CVPR19_PSENet} &CVPR'19&MLT &86.92 &84.50 &85.69 &1.6\\
			LSAE \cite{CVPR19_Embedding}  &CVPR'19&Syn&85.0 &88.3 &86.6 &3.0\\
			CRAFT \cite{CRAFT}  &CVPR'19&Syn&84.3 &89.8  &86.9 &-\\
			LOMO \cite{CVPR19_LOMO} &CVPR'19 &MLT+ &83.5 &\textbf{91.3} &87.2 &3.4\\
			ATRR \cite{CVPR19_ATRR} &CVPR'19&-  &86.0 &89.2 &\textbf{87.6} &-\\
			DRRG \cite{CVPR2020_DRRG}  &CVPR'20&MLT&84.69 &88.53 &86.56 & -\\
			ContourNet \cite{CVPR2020_ContourNet}  &CVPR'20&-&86.1 &87.6 &86.9 &3.5\\
			DB \cite{DB}  &AAAI'20&Syn &83.2 &91.8 &87.3 &12\\
			\hline
			
			\hline
			\textbf{KPN-1280} &-&MLT&83.15 &84.08 &83.61 &12.2 \\
			\textbf{KPN-1920} &-&MLT&84.83 &88.28 &86.52&6.28\\
			\textbf{KPN MS} &-&MLT&\textbf{86.96} &87.84 &87.40&-\\
			\hline
		\end{tabular}
	\end{center}%
\end{table}
\begin{figure*}[htbp]
	\subfigcapskip=0pt
	\centering
	\subfigure[PSENet]{
		\begin{minipage}[b]{0.33\linewidth}
			\centering
			\includegraphics[width=5.9cm,height=10.8cm]{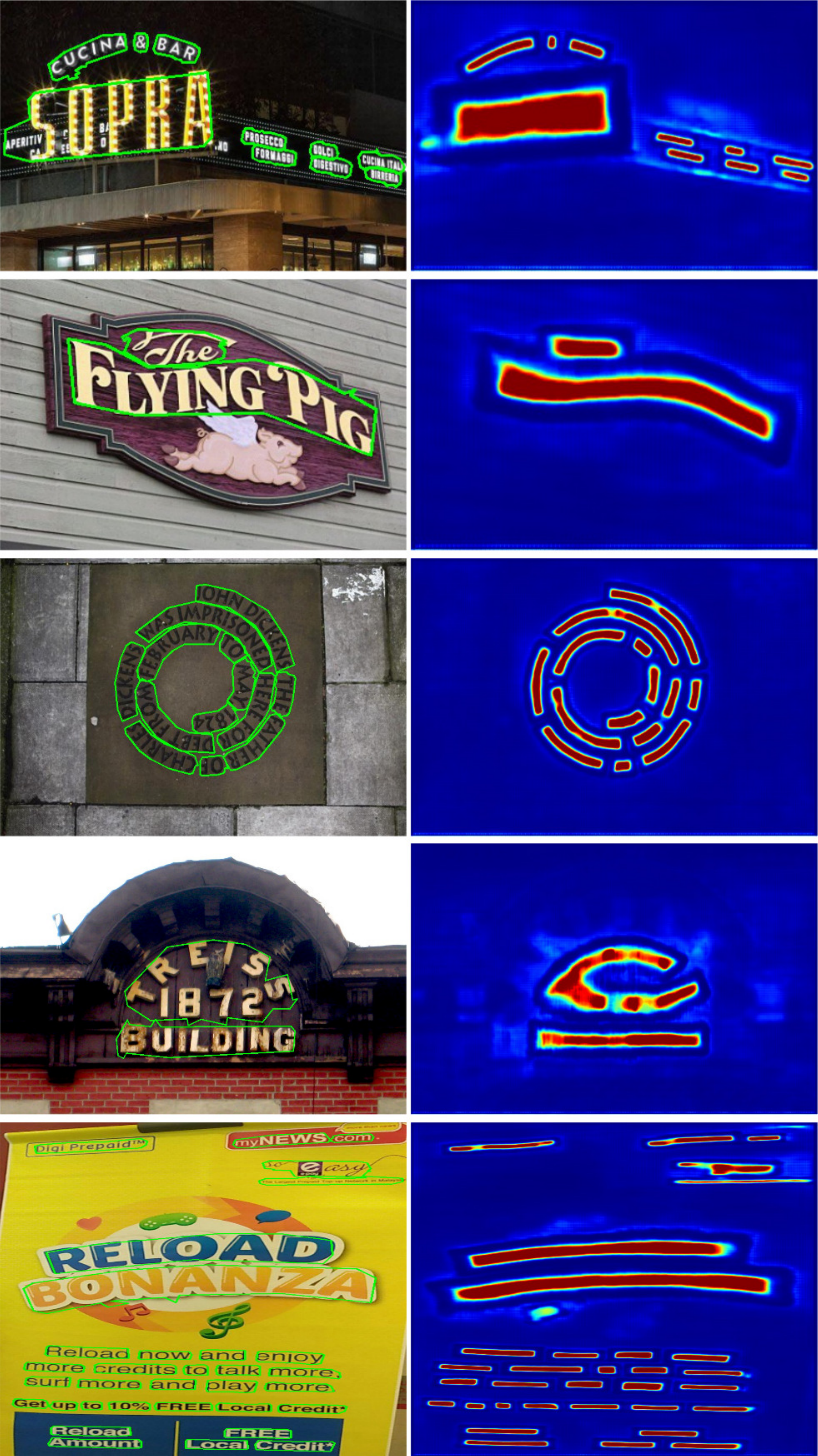}
		\end{minipage}%
	}%
	\subfigure[KPN]{
		\begin{minipage}[b]{0.33\linewidth}
			\centering
			\includegraphics[width=5.9cm,height=10.8cm]{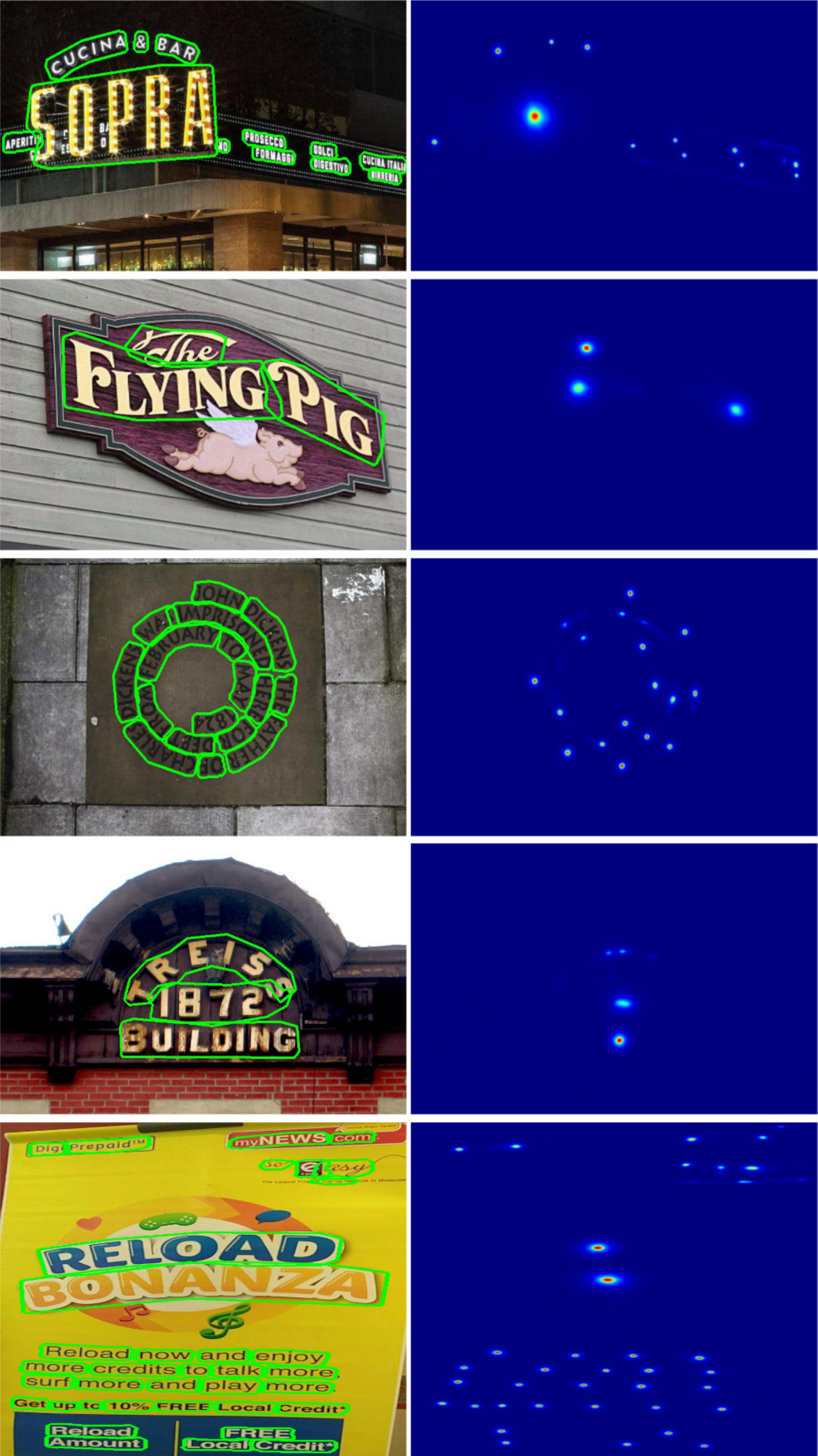}
		\end{minipage}%
	}%
	\subfigure[DB]{
		\begin{minipage}[b]{0.33\linewidth}
			\centering
			\includegraphics[width=5.9cm,height=10.8cm]{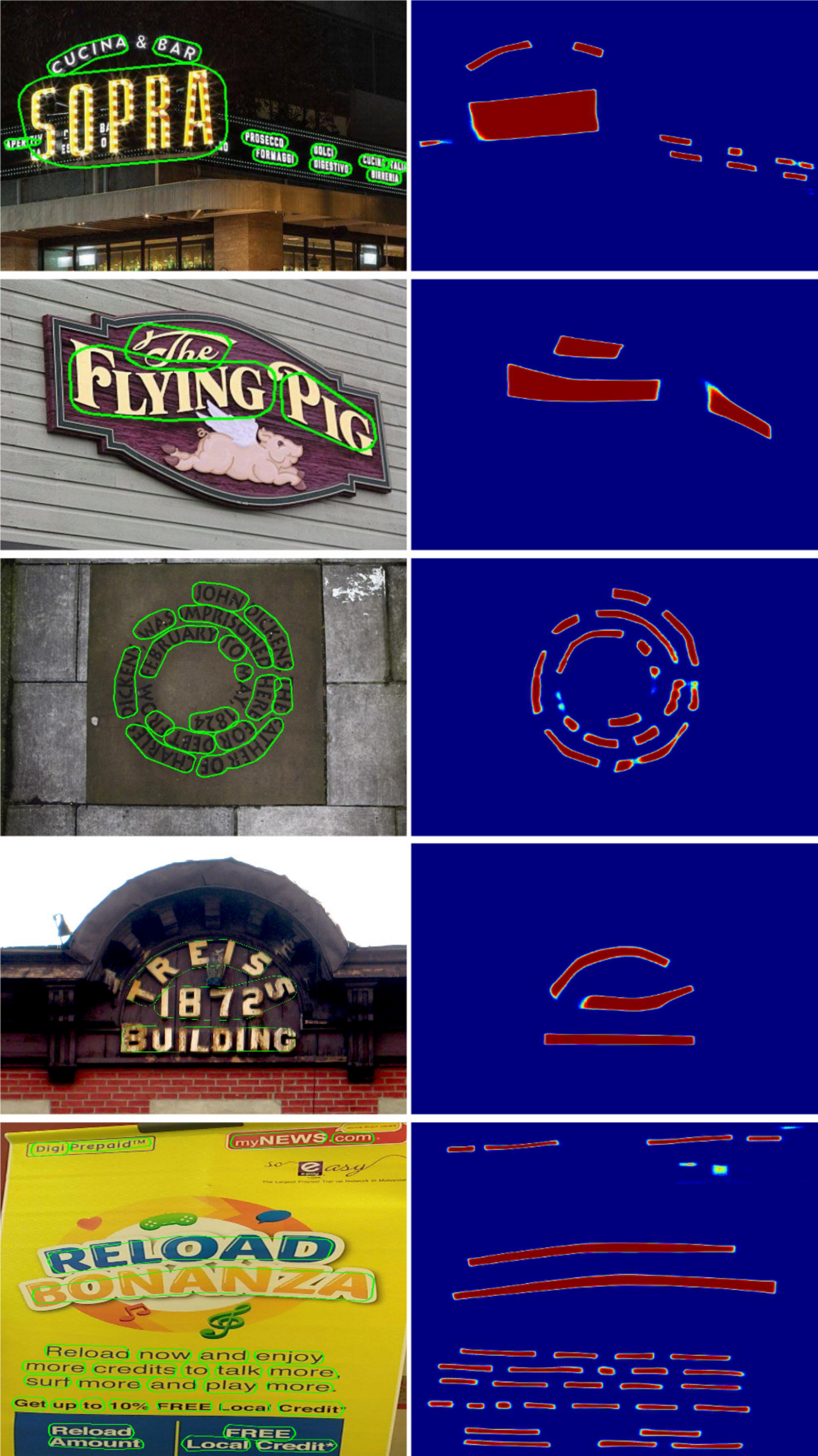}	
		\end{minipage}%
	}%
	\centering
	\caption{ Some visual comparison results with PSENet~\cite{CVPR19_PSENet} and DB~\cite{DB}, where the results of PSENet and DB are reproduced by their official open-source code and model, where DB uses the ResNet-50 with deformable convolution as Bockone and PSENet uses the ResNet-50 as Bockone.}
	\label{fig:compare}
\end{figure*}

\medskip
\subsubsection{Curve Text}
\textbf{Total-Text} mainly contains images with word-level annotated curve texts. We set the threshold $T_c=0.2$, $T_i=0.6$,  and we test two pairs of size $[512, 640]$, $[512, 832]$, \ie, KPN-640, KPN-832. The representative visual results are shown in Fig. \ref{fig:experiment} (a).  As shown in Fig. \ref{fig:experiment} (a), we can see that the predicted results of text instances are well separated. We utilize the official evaluating script in Total-Text \cite{totaltext}, the results are listed in Table \ref{table:TotalText}. Our KPN outperforms DRRG \cite{CVPR2020_DRRG} by 1.38\% in terms of  H-mean, meanwhile outperforms ContourNet by 11.23 FPS in speed. PAN proposes a new network that is faster than ResNet-50, while DB-640 only contains an individual FCN branch. However, our KPN respectively outperforms PAN-640 and DB by 2.11\% and 2.41\% in terms of H-means. In addition, KPN achieves comparable efficiency with PAN and DB-640. Overall, our KPN achieves state-of-the-art performance on Total-Text.

\begin{figure}[htbp]
	\subfigcapskip=0pt
	\centering
	\subfigure[Detected contour]{
		\begin{minipage}[b]{0.325\linewidth}
			\centering
			\includegraphics[width=2.95cm,height=2.2cm]{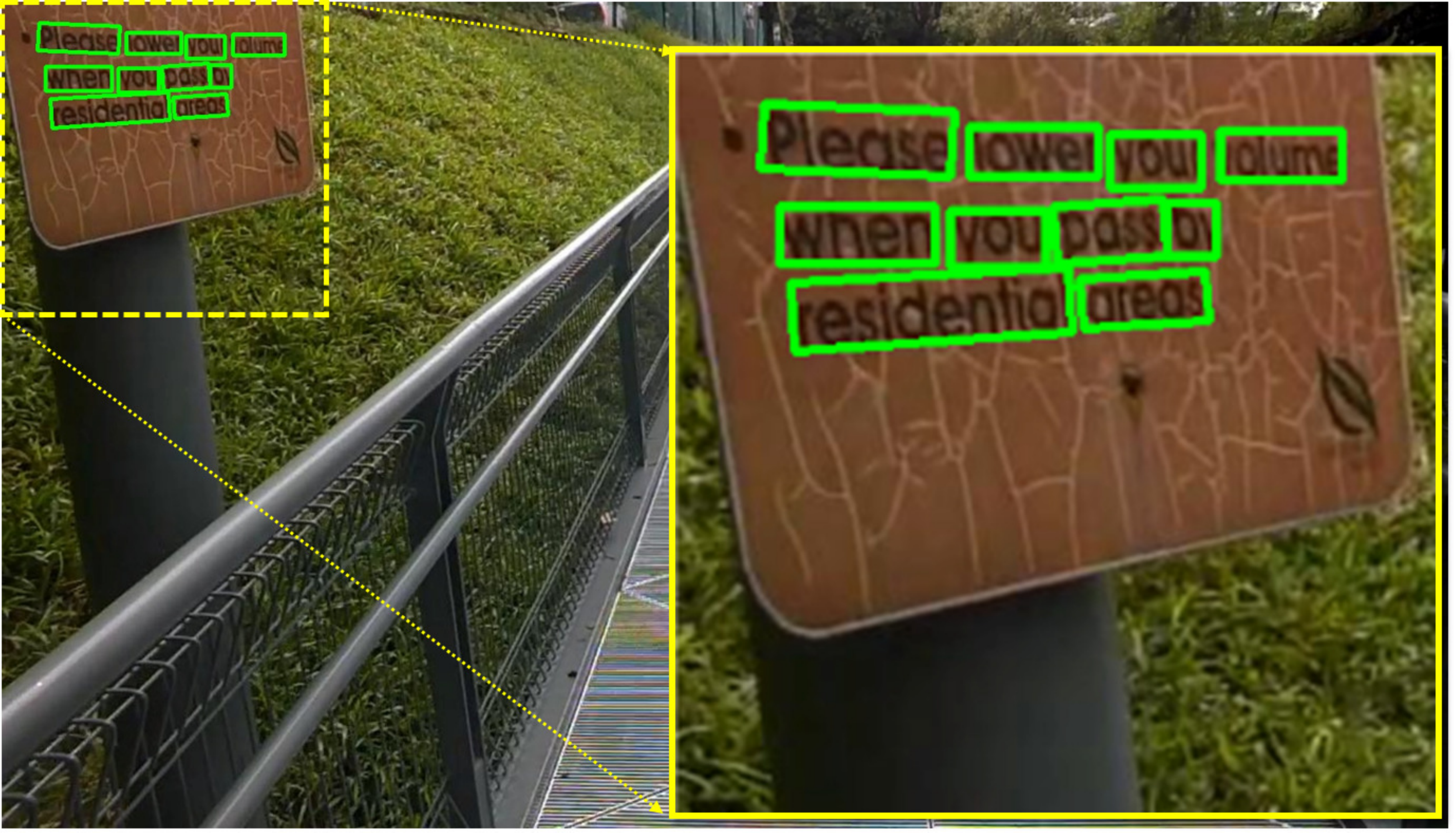}\\
		\end{minipage}%
	}%
	\subfigure[Gaussian center map]{
		\begin{minipage}[b]{0.325\linewidth}
			\centering
			\centering
			\includegraphics[width=2.95cm,height=2.2cm]{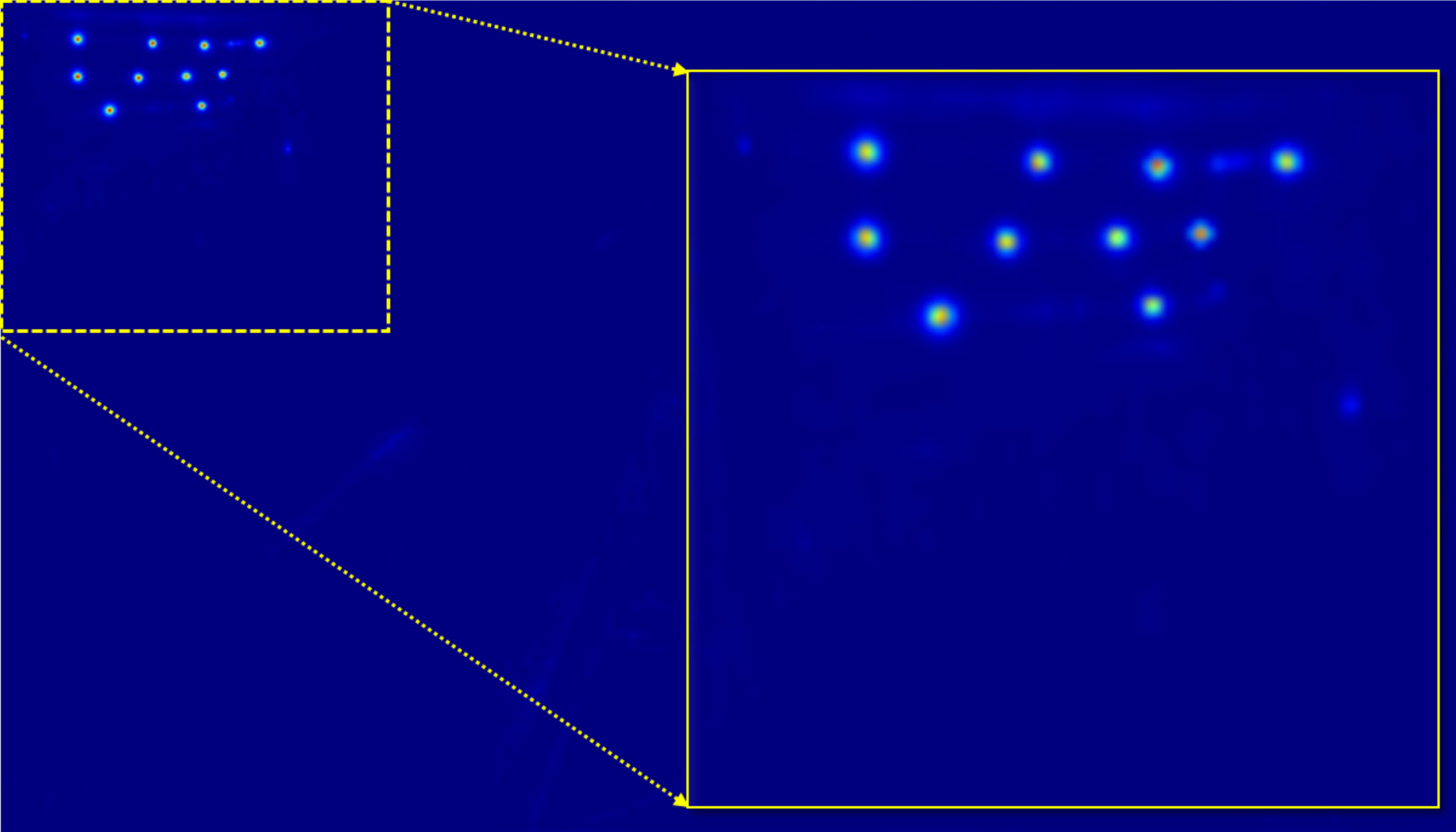}\\
		\end{minipage}%
	}%
	\subfigure[$ S $]{
		\begin{minipage}[b]{0.325\linewidth}
			\centering
			\centering
			\includegraphics[width=2.95cm,height=2.2cm]{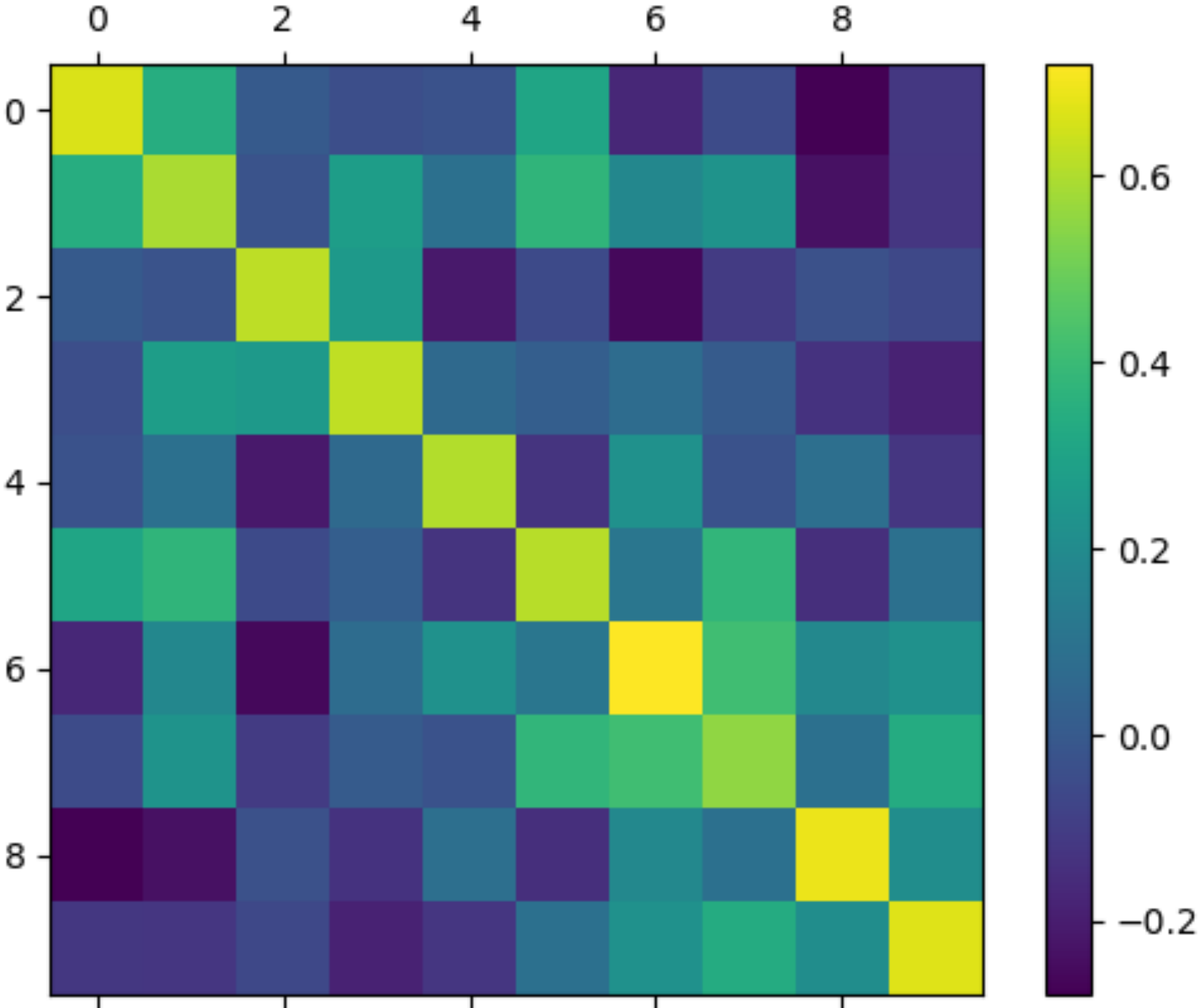}\\
		\end{minipage}%
	}%
	\centering
	\caption{ Visualizing  the similarity matrix $ S $ for Fig~\ref{fig:experiment}(c). Text instances are tiny and dense in this image}
	\label{fig:kps_exp_ic15}
	\vspace{-1.0em}
\end{figure}

\textbf{CTW-1500} mainly contains images with curve texts with line-level annotations. We set the threshold $T_c=0.2$, $T_i=0.625$, and and test in the size $[512, 640]$, $[512, 832]$, \ie, KPN-640, KPN-832. Since CTW-1500 is annotated with line-level, it may be difficult to predict the center point of the text line. Thus our KPN may fail in detecting very long text. The detailed experimental results are listed in able \ref{table:CTW1500}. According to  Table \ref{table:CTW1500}, our KPN outperforms ContourNet \cite{CVPR2020_ContourNet} by 1.32\% in terms of H-means, meanwhile surpassing it on efficiency with a significant margin (KPN 16.30 FPS vs. ContourNet 4.5 FPS). And our KPN is comparable with DRRG \cite{CVPR2020_DRRG} (84.27\% vs. 84.45\%). Apparently, our KPN achieves promising performance on CTW-1500.

According to the experimental results on Total-Text and CTW-1500, KPN achieves promising efficiency and superior performance. The efficiency lies in our post-processing-free segmentation-based framework that cleverly avoids computational-cost post-processing.
In addition, as shown in Fig.~\ref{fig:experiment} (a) and (b), KPN effectively separates neighboring text instances by classifying different texts into instance-independent feature maps. The effectiveness lies in the orthogonality between kernel proposals. When the kernel proposals are orthogonal between each other, they mainly have important self-information learned by the network and position information by position embedding instead of the shared information of other texts.

\subsubsection{Quadrilateral Text}
\textbf{ICDAR 2015} mainly contains quadrilateral texts annotated with word-level. Let $T_c=0.24$, $T_i=0.8$, and we resize the longest side to 1,280 and 1,920 (\ie, KPN-1280, KPN-1920). ICDAR 2015 contains a lot of small texts, but it may be hard to precisely detect the masks of small texts, especially utilizing 1/2 scale of feature maps of the input image. The detailed experimental results are listed in Table \ref{table:tbICDAR15}. According to  Table \ref{table:tbICDAR15}, our KPN outperforms PAN \cite{PAN} by 3.62\% in terms of H-means, outperforms TextDragon~\cite{TextDragon} by 3.47\% in terms of H-means. In addition, representative visual results are shown in Fig.~\ref{fig:experiment}(c). According to the experimental results on quadrilateral texts of ICDAR 2015, we can find that the predictions of independent text masks are not very satisfactory on tiny texts, as shown in Fig. 10(c). Hence, we further analyze the similarity matrix $ S $ of Fig.~\ref{fig:experiment} (c) in Fig.~\ref{fig:kps_exp_ic15}. From Fig.~\ref{fig:kps_exp_ic15} (c), we can find that the similarity matrix $ S $ is not orthogonal enough, inducing some neighboring text instances with small response values of center points. Fortunately, these small response values can be easily filtered by a threshold ($ T_{i} = 0.8 $). Hence, our method can still accurately separate adjacent texts in this case.

\begin{figure}[htbp]
	\subfigcapskip=-4pt
	\centering
	\subfigure[]{
		\begin{minipage}[t]{0.49\linewidth}
			\centering
			\includegraphics[width=4.35cm,height=9.0cm]{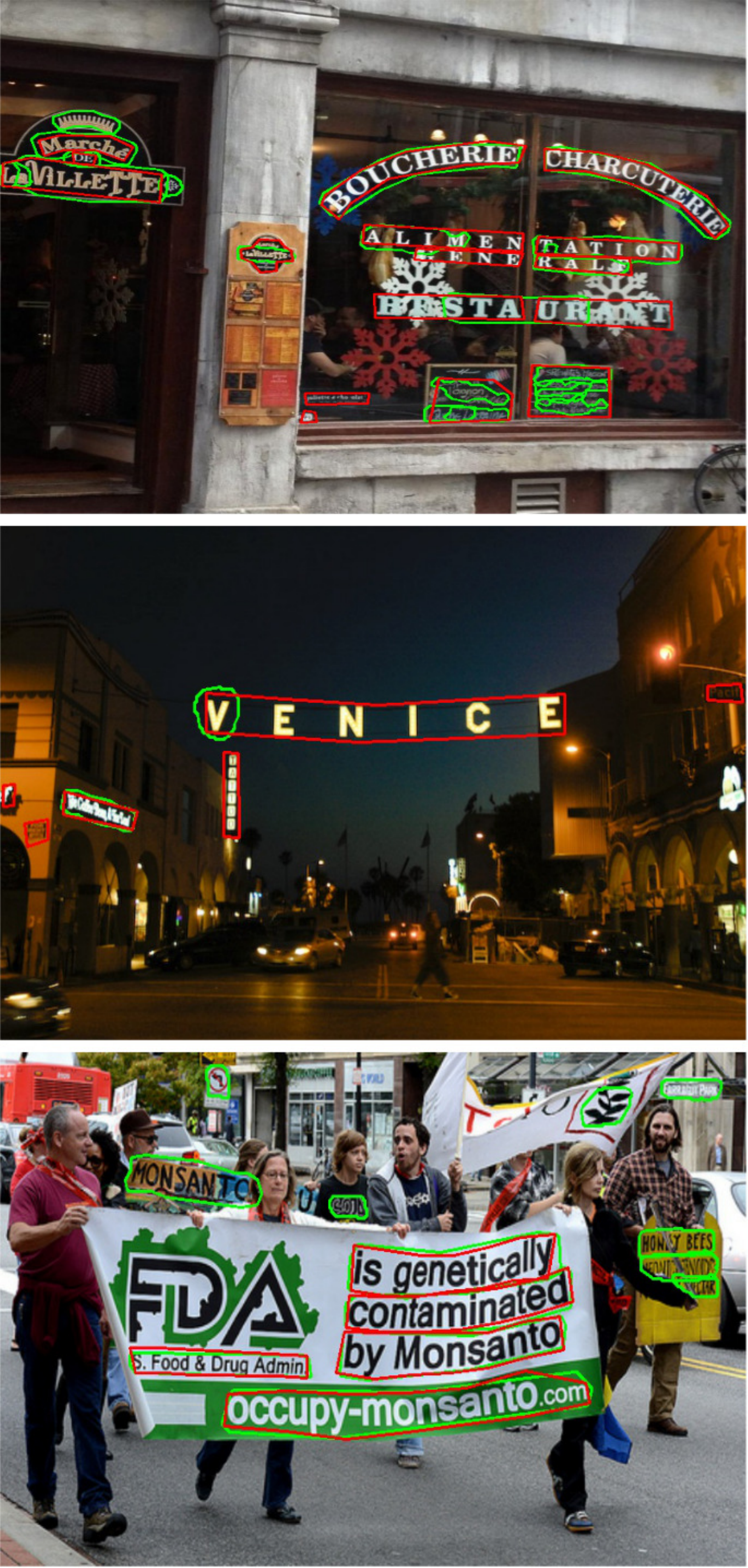}
		\end{minipage}%
	}%
	\subfigure[]{
		\begin{minipage}[t]{0.49\linewidth}
			\centering
			\includegraphics[width=4.35cm,height=9.0cm]{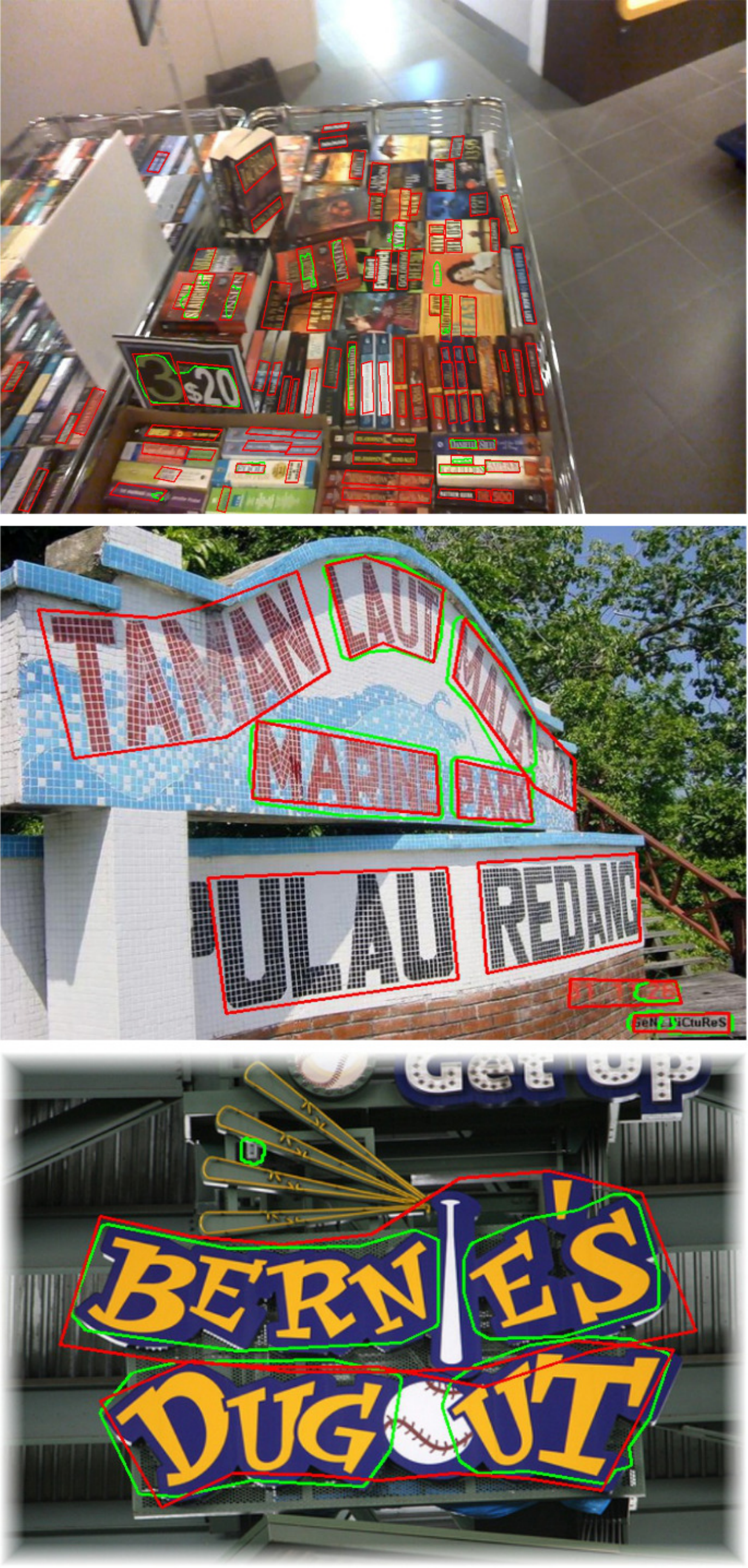}
		\end{minipage}%
	}%
	\centering
	\caption{ Some visual examples of failure cases. Objects included in green contours are results of KPN; objects included in red contours are ground-truths.}
	\label{fig:weakness}
\end{figure}

\subsubsection{Visual Comparison} As shown in Fig.~\ref{fig:compare}, we give a more intuitive comparison between the presented method and the state-of-the-art methods (including PSENet~\cite{CVPR19_PSENet} and DB~\cite{DB}) by
intermediate visual comparison. In Fig.~\ref{fig:compare}, we visualize the final detections, the Gaussian center in KPN, the min text kernel in PSENet, and the probability map in DB.
PSENet uses the min text kernel to separate neighboring text and then adopts Scale Expansion Algorithm to reconstruct text instances, which is not efficient and also fails frequently in many cases. Therefore, the detection speed and performance of PSENet are not so satisfactory. DB uses the probability map which is obtained by shrinking the annotations with Vatti clipping algorithm~\cite{poly_clip} to separate neighboring text. Then, it also uses the inverse transformation of the Vatti clipping algorithm~\cite{poly_clip} to get the final detection boundaries. Therefore, the detection speed of DB is quite fast. However, due to the manual scaling rules, the final detection boundaries of DB can not cover the text well in many cases, either too large or too small, as shown in Fig.~\ref{fig:compare} (c). Although it does not affect the performance of evaluation, inaccurate detection boundaries will lead to various problems in application, such as the recognition model can not correctly recognize characters in the OCR system. In comparison, our method can not only separate neighboring texts well, but also ensure accurate boundary detection and fast detection speed, as shown in Fig.~\ref{fig:compare} (b) and listed in Table~\ref{table:TotalText}\&\ref{table:CTW1500}.

\subsection{Weakness}
As demonstrated in the previous experiments, KPN achieves superior performance in detecting texts of arbitrary shapes. But there is still a chance of failure cases due to the complexity of scene images, such as object occlusion, large character spacing. Some failure examples are shown in Fig.~\ref{fig:weakness}. As shown in Fig.~\ref{fig:weakness}, KPN has some false detections on some text-like areas and miss detections for some extremely large or small complex text. These cases are still very challenging and nontrivial in the paradigm of text detection. However, KPN can successfully detect some text instances with missing annotations, as shown in the bottom image of Fig.~\ref{fig:weakness}(a).

\section{Conclusion} \label{Conclusion}
In this paper, we propose an innovative Kernel Proposal Network for arbitrary shape text detection. The proposed KPN is the first to introduce the dynamic convolution kernel strategy to efficiently and effectively separate neighboring text instances by classifying different texts into instance-independent feature maps. Our KPN is efficient and does not rely on complex post-processing. In addition, we also propose a novel orthogonal learning loss (OLL) that directly enforces the independence between kernel proposals via orthogonal constraints. Specifically, our kernel proposals contain important self-information and position information, resulting in the effectiveness of separating neighboring texts and improving the robustness against tiny intervals or unclear boundaries. Extensive experiments on challenging datasets verify the impressive performance and efficiency of our method.

\section*{Acknowledgment}
This research was supported by the National Key Research and Development Program of China (2020AAA09701), National Science Fund for Distinguished Young Scholars (62125601), National Natural Science Foundation of China
(62076024, 62172035, 62006018, 61806017).

\bibliographystyle{IEEEtran}
\bibliography{IEEEbib}

\begin{thebibliography}{10}
\providecommand{\url}[1]{#1}
\csname url@samestyle\endcsname
\providecommand{\newblock}{\relax}
\providecommand{\bibinfo}[2]{#2}
\providecommand{\BIBentrySTDinterwordspacing}{\spaceskip=0pt\relax}
\providecommand{\BIBentryALTinterwordstretchfactor}{4}
\providecommand{\BIBentryALTinterwordspacing}{\spaceskip=\fontdimen2\font plus
\BIBentryALTinterwordstretchfactor\fontdimen3\font minus
  \fontdimen4\font\relax}
\providecommand{\BIBforeignlanguage}[2]{{%
\expandafter\ifx\csname l@#1\endcsname\relax
\typeout{** WARNING: IEEEtran.bst: No hyphenation pattern has been}%
\typeout{** loaded for the language `#1'. Using the pattern for}%
\typeout{** the default language instead.}%
\else
\language=\csname l@#1\endcsname
\fi
#2}}
\providecommand{\BIBdecl}{\relax}
\BIBdecl

\bibitem{Faster-rcnn}
S.~Ren, K.~He, R.~B. Girshick, and J.~Sun, ``Faster {R-CNN:} {Towards}
  real-time object detection with region proposal networks,'' \emph{{IEEE}
  Trans. Pattern Anal. Mach. Intell.}, vol.~39, no.~6, pp. 1137--1149, 2017.

\bibitem{FPN}
T.~Lin, P.~Doll{\'{a}}r, R.~B. Girshick, K.~He, B.~Hariharan, and S.~J.
  Belongie, ``Feature pyramid networks for object detection,'' in
  \emph{{CVPR}}, 2017, pp. 936--944.

\bibitem{DABNet}
Q.~He, X.~Sun, Z.~Yan, and K.~Fu, ``Dabnet: Deformable contextual and
  boundary-weighted network for cloud detection in remote sensing images,''
  \emph{IEEE Transactions on Geoscience and Remote Sensing}, pp. 1--16, 2021.

\bibitem{FCN}
J.~Long, E.~Shelhamer, and T.~Darrell, ``Fully convolutional networks for
  semantic segmentation,'' in \emph{CVPR}, 2015, pp. 3431--3440.

\bibitem{U-Net}
O.~Ronneberger, P.~Fischer, and T.~Brox, ``U-net: Convolutional networks for
  biomedical image segmentation,'' in \emph{MICCAI}, N.~Navab, J.~Hornegger,
  W.~M.~W. III, and A.~F. Frangi, Eds., vol. 9351, 2015, pp. 234--241.

\bibitem{EAST}
X.~Zhou, C.~Yao, H.~Wen, Y.~Wang, S.~Zhou, W.~He, and J.~Liang, ``{EAST:} {An}
  efficient and accurate scene text detector,'' in \emph{{CVPR}}, 2017, pp.
  2642--2651.

\bibitem{RRPN}
J.~Ma, W.~Shao, H.~Ye, L.~Wang, H.~Wang, and Y.~Z. anda Xiangyang~Xue,
  ``Arbitrary-oriented scene text detection via rotation proposals,''
  \emph{{IEEE} Trans. Multimedia}, vol.~20, no.~11, pp. 3111--3122, 2018.

\bibitem{CVPR2020_DRRG}
S.-X. Zhang, X.~Zhu, J.~Hou, C.~Liu, C.~Yang, H.~Wang, and X.~Yin, ``Deep
  relational reasoning graph network for arbitrary shape text detection,'' in
  \emph{{CVPR}}, 2020, pp. 9696--9705.

\bibitem{PixelLink}
D.~Deng, H.~Liu, X.~Li, and D.~Cai, ``{PixelLink}: Detecting scene text via
  instance segmentation,'' in \emph{AAAI}, 2018, pp. 6773--6780.

\bibitem{PAN}
W.~Wang, E.~Xie, X.~Song, Y.~Zang, W.~Wang, T.~Lu, G.~Yu, and C.~Shen,
  ``Efficient and accurate arbitrary-shaped text detection with pixel
  aggregation network,'' in \emph{{ICCV}}, 2019, pp. 8439--8448.

\bibitem{CVPR19_Embedding}
Z.~Tian, M.~Shu, P.~Lyu, R.~Li, C.~Zhou, X.~Shen, and J.~Jia, ``Learning
  shape-aware embedding for scene text detection,'' in \emph{CVPR}, 2019, pp.
  4234--4243.

\bibitem{DB}
M.~Liao, Z.~Wan, C.~Yao, K.~Chen, and X.~Bai, ``Real-time scene text detection
  with differentiable binarization,'' in \emph{{AAAI}}, 2020, pp.
  11\,474--11\,481.

\bibitem{CVPR19_PSENet}
W.~Wang, E.~Xie, X.~Li, W.~Hou, T.~Lu, G.~Yu, and S.~Shao, ``Shape robust text
  detection with progressive scale expansion network,'' in \emph{CVPR}, 2019,
  pp. 9336--9345.

\bibitem{TextField}
Y.~Xu, Y.~Wang, W.~Zhou, Y.~Wang, Z.~Yang, and X.~Bai, ``Textfield: Learning a
  deep direction field for irregular scene text detection,'' \emph{{IEEE}
  Trans. Image Processing}, vol.~28, no.~11, pp. 5566--5579, 2019.

\bibitem{poly_clip}
B.~R. Vatti, ``A generic solution to polygon clipping,'' \emph{Commun. {ACM}},
  vol.~35, no.~7, pp. 56--63, 1992.

\bibitem{Maskrcnn}
K.~He, G.~Gkioxari, P.~Doll{\'{a}}r, and R.~B. Girshick, ``Mask {R-CNN},'' in
  \emph{{ICCV}}, 2017, pp. 2980--2988.

\bibitem{FTSN}
Y.~Dai, Z.~Huang, Y.~Gao, Y.~Xu, K.~Chen, J.~Guo, and W.~Qiu, ``Fused text
  segmentation networks for multi-oriented scene text detection,'' in
  \emph{ICPR}, 2018, pp. 3604--3609.

\bibitem{CVPR19_LOMO}
C.~Zhang, B.~Liang, Z.~Huang, M.~En, J.~Han, E.~Ding, and X.~Ding, ``Look more
  than once: An accurate detector for text of arbitrary shapes,'' in
  \emph{CVPR}, 2019, pp. 10\,552--10\,561.

\bibitem{Mask_TextSpotter_v3}
M.~Liao, G.~Pang, J.~Huang, T.~Hassner, and X.~Bai, ``Mask textspotter v3:
  Segmentation proposal network for robust scene text spotting,'' \emph{CoRR},
  vol. abs/2007.09482, 2020.

\bibitem{textboxes}
M.~Liao, B.~Shi, X.~Bai, X.~Wang, and W.~Liu, ``Textboxes: {A} fast text
  detector with a single deep neural network,'' in \emph{{AAAI}}, 2017, pp.
  4161--4167.

\bibitem{textboxes++}
M.~Liao, B.~Shi, and X.~Bai, ``Textboxes++: A single-shot oriented scene text
  detector,'' \emph{{IEEE} Trans. Image Processing}, vol.~27, no.~8, pp.
  3676--3690, 2018.

\bibitem{DDR}
W.~He, X.-Y. Zhang, F.~Yin, and C.-L. Liu, ``Deep direct regression for
  multi-oriented scene text detection,'' in \emph{ICCV}, 2017, pp. 745--753.

\bibitem{IOU}
J.~Yu, Y.~Jiang, Z.~Wang, Z.~Cao, and T.~S. Huang, ``Unitbox: An advanced
  object detection network,'' in \emph{{ACM MM}}, 2016, pp. 516--520.

\bibitem{HAM}
J.~Hou, X.~Zhu, C.~Liu, K.~Sheng, L.~Wu, H.~Wang, and X.~Yin, ``{HAM:} hidden
  anchor mechanism for scene text detection,'' \emph{{IEEE} Trans. Image
  Process.}, vol.~29, pp. 7904--7916, 2020.

\bibitem{PBNet}
X.~Sun, P.~Wang, C.~Wang, Y.~Liu, and K.~Fu, ``Pbnet: Part-based convolutional
  neural network for complex composite object detection in remote sensing
  imagery,'' \emph{ISPRS Journal of Photogrammetry and Remote Sensing}, vol.
  173, pp. 50--65, 2021.

\bibitem{CTPN}
Z.~Tian, W.~Huang, T.~He, P.~He, and Y.~Qiao, ``Detecting text in natural image
  with connectionist text proposal network,'' in \emph{ECCV}, 2016, pp. 56--72.

\bibitem{SegLink}
B.~Shi, X.~Bai, and S.~J. Belongie, ``Detecting oriented text in natural images
  by linking segments,'' in \emph{CVPR}, 2017, pp. 3482--3490.

\bibitem{SegLink++}
J.~Tang, Z.~Yang, Y.~Wang, Q.~Zheng, Y.~Xu, and X.~Bai, ``Seglink++: Detecting
  dense and arbitrary-shaped scene text by instance-aware component grouping,''
  \emph{Pattern Recognition}, vol.~96, 2019.

\bibitem{CRAFT}
Y.~Baek, B.~Lee, D.~Han, S.~Yun, and H.~Lee, ``Character region awareness for
  text detection,'' in \emph{CVPR}, 2019, pp. 9365--9374.

\bibitem{TextDragon}
W.~Feng, W.~He, F.~Yin, X.~Zhang, and C.~Liu, ``Textdragon: An end-to-end
  framework for arbitrary shaped text spotting,'' in \emph{{ICCV}}, 2019, pp.
  9075--9084.

\bibitem{CVPR19_ATRR}
X.~Wang, Y.~Jiang, Z.~Luo, C.-L. Liu, H.~Choi, and S.~Kim, ``Arbitrary shape
  scene text detection with adaptive text region representation,'' in
  \emph{CVPR}, 2019, pp. 6449--6458.

\bibitem{ContourNet}
Y.~Wang, H.~Xie, Z.-J. Zha, M.~Xing, Z.~Fu, and Y.~Zhang, ``Contournet: Taking
  a further step toward accurate arbitrary-shaped scene text detection,'' in
  \emph{CVPR}, 2020, pp. 11\,753--11\,762.

\bibitem{ABCNet}
Y.~Liu, H.~Chen, C.~Shen, T.~He, L.~Jin, and L.~Wang, ``Abcnet: Real-time scene
  text spotting with adaptive bezier-curve network,'' in \emph{CVPR}, 2020, pp.
  9806--9815.

\bibitem{FCENet}
Y.~Zhu, J.~Chen, L.~Liang, Z.~Kuang, L.~Jin, and W.~Zhang, ``Fourier contour
  embedding for arbitrary-shaped text detection,'' in \emph{CVPR}, 2021, pp.
  3123--3131.

\bibitem{TextRay}
F.~Wang, Y.~Chen, F.~Wu, and X.~Li, ``Textray: Contour-based geometric modeling
  for arbitrary-shaped scene text detection,'' in \emph{ACM-MM}, C.~W. Chen,
  R.~Cucchiara, X.~Hua, G.~Qi, E.~Ricci, Z.~Zhang, and R.~Zimmermann, Eds.,
  2020, pp. 111--119.

\bibitem{PCR}
P.~Dai, S.~Zhang, H.~Zhang, and X.~Cao, ``Progressive contour regression for
  arbitrary-shape scene text detection,'' in \emph{CVPR}, 2021, pp. 7393--7402.

\bibitem{TextBPN}
S.-X. Zhang, X.~Zhu, C.~Yang, H.~Wang, and X.-C. Yin, ``Adaptive boundary
  proposal network for arbitrary shape text detection,'' in \emph{ICCV},
  October 2021, pp. 1305--1314.

\bibitem{IncepText}
Q.~Yang, M.~Cheng, W.~Zhou, Y.~Chen, M.~Qiu, and W.~Lin, ``Inceptext: A new
  inception-text module with deformable psroi pooling for multi-oriented scene
  text detection,'' in \emph{{IJCAI}}, 2018, pp. 1071--1077.

\bibitem{AdaptIS}
K.~Sofiiuk, O.~Barinova, and A.~Konushin, ``Adaptis: Adaptive instance
  selection network,'' in \emph{{ICCV}}, 2019, pp. 7354--7362.

\bibitem{AdaIN}
T.~Karras, S.~Laine, and T.~Aila, ``A style-based generator architecture for
  generative adversarial networks,'' in \emph{{CVPR}}, 2019, pp. 4401--4410.

\bibitem{FCOS}
Z.~Tian, C.~Shen, H.~Chen, and T.~He, ``Fcos: Fully convolutional one-stage
  object detection,'' pp. 9626--9635, 2019.

\bibitem{condinst}
Z.~Tian, C.~Shen, and H.~Chen, ``Conditional convolutions for instance
  segmentation,'' in \emph{{ECCV}}, ser. Lecture Notes in Computer Science,
  A.~Vedaldi, H.~Bischof, T.~Brox, and J.~Frahm, Eds., vol. 12346, 2020, pp.
  282--298.

\bibitem{solov2}
X.~Wang, R.~Zhang, T.~Kong, L.~Li, and C.~Shen, ``Solov2: Dynamic, faster and
  stronger,'' \emph{arXiv preprint arXiv:2003.10152}, 2020.

\bibitem{TextSnake}
S.~Long, J.~Ruan, W.~Zhang, X.~He, W.~Wu, and C.~Yao, ``Textsnake: {A} flexible
  representation for detecting text of arbitrary shapes,'' in \emph{{ECCV}},
  2018, pp. 19--35.

\bibitem{cornernet}
H.~Law and J.~Deng, ``Cornernet: Detecting objects as paired keypoints,'' in
  \emph{{ECCV}}, V.~Ferrari, M.~Hebert, C.~Sminchisescu, and Y.~Weiss, Eds.,
  vol. 11218, 2018, pp. 765--781.

\bibitem{center_point}
K.~Duan, S.~Bai, L.~Xie, H.~Qi, Q.~Huang, and Q.~Tian, ``Centernet: Keypoint
  triplets for object detection,'' in \emph{{ICCV}}, 2019, pp. 6568--6577.

\bibitem{ResNet}
K.~He, X.~Zhang, S.~Ren, and J.~Sun, ``Deep residual learning for image
  recognition,'' in \emph{{CVPR}}, 2016, pp. 770--778.

\bibitem{semi-conv}
D.~Novotn{\'{y}}, S.~Albanie, D.~Larlus, and A.~Vedaldi, ``Semi-convolutional
  operators for instance segmentation,'' in \emph{{ECCV}}, ser. Lecture Notes
  in Computer Science, V.~Ferrari, M.~Hebert, C.~Sminchisescu, and Y.~Weiss,
  Eds., vol. 11205, 2018, pp. 89--105.

\bibitem{coordconv}
R.~Liu, J.~Lehman, P.~Molino, F.~P. Such, E.~Frank, A.~Sergeev, and
  J.~Yosinski, ``An intriguing failing of convolutional neural networks and the
  coordconv solution,'' in \emph{NeurIPS}, 2018, pp. 9628--9639.

\bibitem{diceloss}
F.~Milletari, N.~Navab, and S.~Ahmadi, ``V-net: Fully convolutional neural
  networks for volumetric medical image segmentation,'' in \emph{3DV}, 2016,
  pp. 565--571.

\bibitem{focalloss}
T.~Lin, P.~Goyal, R.~B. Girshick, K.~He, and P.~Doll{\'{a}}r, ``Focal loss for
  dense object detection,'' in \emph{{ICCV}}, 2017, pp. 2999--3007.

\bibitem{OHEM}
A.~Shrivastava, A.~Gupta, and R.~B. Girshick, ``Training region-based object
  detectors with online hard example mining,'' in \emph{{CVPR}}, 2016, pp.
  761--769.

\bibitem{totaltext}
C.~K. Chng and C.~S. Chan, ``Total-text: {A} comprehensive dataset for scene
  text detection and recognition,'' in \emph{ICDAR}, 2017, pp. 935--942.

\bibitem{ctw1500}
Y.~Liu, L.~Jin, S.~Zhang, and S.~Zhang, ``Detecting curve text in the wild: New
  dataset and new solution,'' \emph{CoRR}, vol. abs/1712.02170, 2017.

\bibitem{IC15}
D.~Karatzas, L.~Gomez{-}Bigorda, A.~Nicolaou, S.~K. Ghosh, and A.~D.~B. et~al.,
  ``{ICDAR} 2015 competition on robust reading,'' in \emph{{ICDAR}}, 2015, pp.
  1156--1160.

\bibitem{ADAM}
D.~P. Kingma and J.~Ba, ``Adam: {A} method for stochastic optimization,'' in
  \emph{ICLR}, 2015.

\bibitem{MLT}
N.~Nayef, F.~Yin, I.~Bizid, H.~Choi, Y.~Feng, D.~Karatzas, and Z.~L. et~al.,
  ``{ICDAR2017} robust reading challenge on multi-lingual scene text detection
  and script identification - {RRC-MLT},'' in \emph{{ICDAR}}, 2017, pp.
  1454--1459.

\bibitem{MSR}
C.~Xue, S.~Lu, and W.~Zhang, ``{MSR:} multi-scale shape regression for scene
  text detection,'' in \emph{IJCAI}, 2019, pp. 989--995.

\bibitem{CVPR19_CSE}
Z.~Liu, G.~Lin, S.~Yang, F.~Liu, W.~Lin, and W.~L. Goh, ``Towards robust curve
  text detection with conditional spatial expansion,'' in \emph{CVPR}, 2019,
  pp. 7269--7278.

\bibitem{CVPR2020_ContourNet}
Y.~Wang, H.~Xie, Z.~Zha, M.~Xing, Z.~Fu, and Y.~Zhang, ``Contournet: Taking a
  further step toward accurate arbitrary-shaped scene text detection,'' in
  \emph{{CVPR}}, 2020, pp. 11\,750--11\,759.

\bibitem{CTD}
Y.~Liu, L.~Jin, S.~Zhang, C.~Luo, and S.~Zhang, ``Curved scene text detection
  via transverse and longitudinal sequence connection,'' \emph{Pattern
  Recognition}, vol.~90, pp. 337--345, 2019.

\bibitem{MCN}
Z.~Liu, G.~Lin, S.~Yang, J.~Feng, W.~Lin, and W.~Ling~Goh, ``Learning markov
  clustering networks for scene text detection,'' in \emph{CVPR}, 2018, pp.
  6936--6944.

\bibitem{FOTS}
X.~Liu, D.~Liang, S.~Yan, D.~Chen, Y.~Qiao, and J.~Yan, ``{FOTS: Fast} oriented
  text spotting with a unified network,'' in \emph{{CVPR}}, 2018, pp.
  5676--5685.

\end{thebibliography}

\end{document}